\documentclass{article} %
\usepackage[final]{colm2026_conference}

\usepackage{microtype}
\usepackage{hyperref}
\usepackage{url}
\usepackage{booktabs}

\usepackage{lineno}

\usepackage{amsmath}
\usepackage{amssymb}
\usepackage{mathtools}
\usepackage{amsthm}

\usepackage{epigraph}

\usepackage{booktabs}
\usepackage{multirow}
\usepackage{array}
\usepackage{colortbl}
\usepackage{xcolor}
\usepackage{caption}
\usepackage{subcaption}
\usepackage{adjustbox}
\usepackage{makecell}
\usepackage{siunitx}
\usepackage{cellspace}
\usepackage{float}
\usepackage{lscape}  %
\definecolor{green2}{rgb}{0.56, 0.93, 0.56}
\definecolor{blue2}{rgb}{0.56, 0.56, 0.93}
\usepackage{amsmath,amssymb}

\definecolor{gap1}{HTML}{63BE7B}  %
\definecolor{gap2}{HTML}{B1D580}  %
\definecolor{gap3}{HTML}{FFEB84}  %
\definecolor{gap4}{HTML}{FDB96A}  %
\definecolor{gap5}{HTML}{F8696B}  %
\definecolor{darkblue}{rgb}{0, 0, 0.5}
\hypersetup{colorlinks=true, citecolor=darkblue, linkcolor=darkblue, urlcolor=darkblue}

\usepackage[capitalize,noabbrev]{cleveref}

\newcommand{\shift}[2]{\ensuremath{#1\!\rightarrow\!#2}}

\title{LLMs Exhibit Significantly Lower Uncertainty \\in Creative Writing Than Professional Writers}

\author{Peiqi Sui \\
Department of English\\
McGill University\\
\texttt{peiqi.sui@mail.mcgill.ca} \\
}

\begin{document}

\ifcolmsubmission
\linenumbers
\fi

\maketitle

\begin{abstract}
We argue that uncertainty is a key and understudied limitation of LLMs' performance in creative writing, which is often characterized as trite and cliché-ridden. Literary theory identifies uncertainty as a necessary condition for creative expression, while current alignment strategies steer models away from uncertain outputs to ensure factuality and reduce hallucination. We formalize this tension by quantifying the \textbf{``uncertainty gap''} between human-authored stories and model-generated continuations. Through a controlled information-theoretic analysis of 28 LLMs on high-quality storytelling datasets, we demonstrate that human writing consistently exhibits significantly higher uncertainty than model outputs. We find that instruction-tuned and reasoning models exacerbate this trend compared to their base counterparts; furthermore, the gap is more pronounced in creative writing than in functional domains, and shows a consistent correlation with writing quality. Achieving human-level creativity requires new uncertainty-aware alignment paradigms that can distinguish between destructive hallucinations and the constructive ambiguity required for literary richness.
\end{abstract}

\setlength{\epigraphwidth}{\textwidth}
\epigraph{
``What quality went to form a Man of Achievement, especially in Literature \dots\ was [the] capab[ility] of being in uncertainties, Mysteries, doubts, without any irritable reaching after fact and reason.''
}{— John Keats, \citeyearpar{keats1817}}

\section{Introduction}

What happens when an automated, industrial-scale form of ``irritable reaching after fact and reason'' attempts to write stories or poems? While large language models (LLMs) exhibit some potential \citep{ippolito2022creative,mirowski2023cowriting}, their performance in creative writing remains distinctively limited compared to other domains. Recent evaluations reveal that current models often produce outputs that are trite, repetitive, overly verbose, and cliché-ridden \citep{chakrabarty2024art,chakrabarty2025salvaged}. Structurally, they struggle to sustain dramatic tension and suspense \citep{tian2024narratives}, preserve the diversity of plot and narrative arcs \citep{padmakumar2024diversity,xu2025echoes}, or maintain long-range narrative coherence \citep{gurung2025learning,pham2025frankentext}. Furthermore, their utility as co-creative systems is hampered by their limited ability to provide helpful feedback \citep{rashkin2025feedback} or serve as reliable evaluators of creativity \citep{lu2025rethinking}. Even when individual outputs appear creative, model assistance has been shown to reduce content diversity \citep{doshi2024generative,anderson2024homogenization} and drive cultural homogenization \citep{bhagat2025tales} when scaled to larger groups, raising fundamental concerns regarding the systemic risks that LLMs present to the landscape of human creativity \citep{sourati2025homogenizing}.

\begin{figure}[!ht]
    \vspace{-0.5cm}
    \makebox[\textwidth][c]{\includegraphics[width=1.05\textwidth]{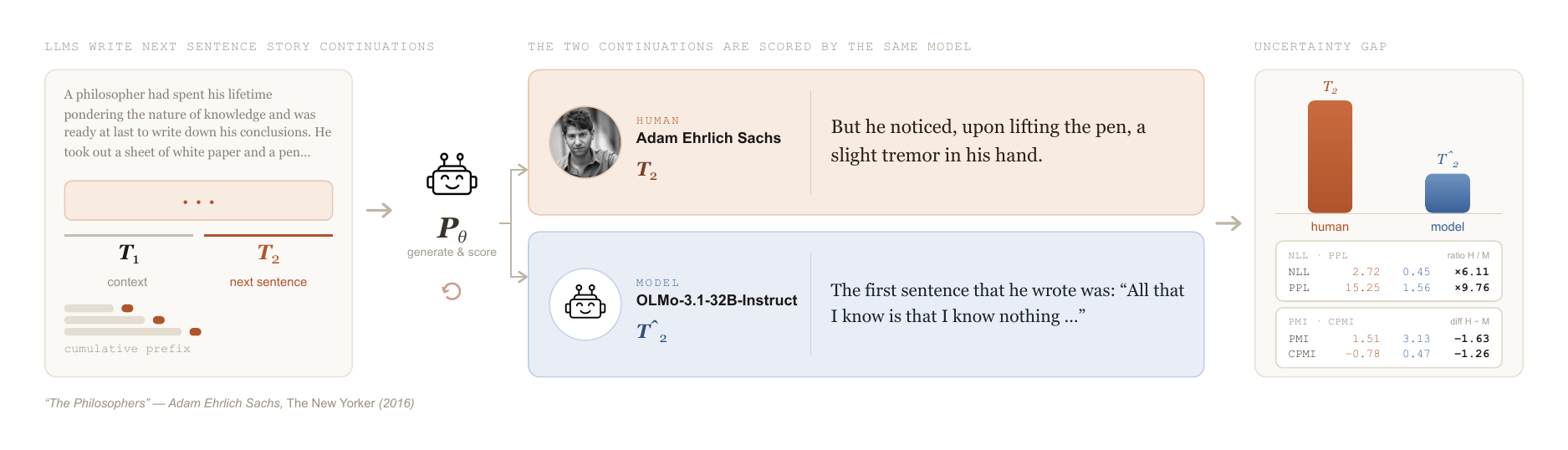}}
    \caption{\textbf{The human--model uncertainty gap.} Given a cumulative prefix $T_1$ from a \textit{New Yorker} story, we compare the author's next sentence $T_2$ against a model-generated continuation $\hat{T}_2$, scoring both under the \emph{same} model (e.g., \textsc{OLMo-3.1-32B-Instruct}). The human continuation presents a plot twist that introduces genuine tension into the story, which is reflected on all uncertainty metrics; the model instead reduces the scene to a stock quotation.}
    \label{fig:uncertainty1}
\end{figure}

One potential factor is that LLMs are often explicitly trained to steer away from uncertain outputs to ensure factuality \citep{kalai2025why}, reduce hallucination \citep{farquhar2024detecting}, and maintain typicality \citep{meister2023locally}. Current alignment strategies aim to reduce ambiguity to ensure instruction-following in general-domain tasks like question-answering \citep{kim2024aligning,li2025condambigqa}, natural language inference \citep{liu2023afraid}, semantic parsing \citep{saparina2025disambiguate}, and coreference resolution \citep{shore2025correct}. Mechanistically, the optimization for reliable adherence to human preferences often forces models into mode collapse, where the probability distribution converges on a narrow set of safe, homogenized attractor states \citep{mohammadi2024creativity,anschel2025group}. This systemic aversion to uncertainty is exacerbated by the KL-divergence regularization terms used in standard alignment algorithms, which penalize the long tail of diverse outputs to prioritize the majority opinion \citep{slocum2025diverse}. Furthermore, reward models used in reinforcement learning from human feedback (RLHF) demonstrate an inherent bias toward verbalized confidence, incentivizing the model to suppress epistemic doubt and simulate certainty even when its internal knowledge is incomplete \citep{leng2025taming}. These pitfalls are particularly acute in creative domains, where aligned LLMs already struggle with reduced output diversity from supervised fine-tuning (SFT) \citep{li2025preserving}, a lack of authorial voice \citep{kreminski2024dearth}, and ''typicality bias''—a systematic preference for predictable patterns learned from human preference data \citep{zhang2025verbalized}.

Meanwhile, literary theory has long posited that uncertainty is a functional requirement for creative expression rather than a defect. Literary richness arises from the simultaneity of conflicting meanings \citep{keats1817}, ambiguity \citep{empson1930ambiguity}, and indeterminacy \citep{derrida1967writing}. Recent scholarship further formalizes failure and linguistic error as key aesthetic categories with unique generative affordances \citep{halberstam2011queer,jones2022glitch,davidson2022distressing}. As such, the literary value of a work is defined by its ''openness'' to multiple interpretations \citep{eco1962open}, transforming the text into a ''writerly'' site of active production where the reader must navigate a field of possibilities \citep{barthes1970sz}. Since meaning-making in literature is collaborative, effective storytelling necessitates ``gaps'' of indeterminacy that elicit the reader's imagination to participate in the story's construction \citep{iser1972reading}. By eliminating these gaps to prioritize unambiguous helpfulness and clarity, current alignment strategies effectively foreclose this co-creative process, optimizing away a key precondition of the literary experience.

This positive view of uncertainty resonates with recent NLP research that shows the usefulness of uncertainty in various research areas like safety \citep{quach2024conformal}, reasoning \citep{hu2024uncertainty}, and alignment \citep{xu2025ask}. Uncertainty-aware methods also show promise in creative contexts by effectively balancing coherence and diversity \citep{garcesarias2024adaptive}. Building in this direction, we draw on literary theory to formalize the link between uncertainty and creativity. We propose an information-theoretic framework to quantify the human–model uncertainty gap, empirically testing whether LLMs' systematic aversion to uncertainty manifests as a measurable distributional divergence from human-authored fiction. While the lack of surprise is highlighted as a key limitation of LLM-generated stories \citep{tian2024narratives,ismayilzada2025evaluating}, few studies ground it in information theory through rigorous uncertainty estimation \citep{garcesarias2025geometry,peeperkorn2025mind}; the closest precedent is \citet{wilmot2020modelling}, which model narrative suspense as uncertainty reduction at the span level, but not as a more general distributional property sharpened by alignment that separates human-written and generated stories.

Is uncertainty empirically correlated with creativity, and does constraining LLMs to less uncertain outputs make them fundamentally bad at creative writing? Our controlled story completion experiment seeks to answer these questions by comparing the uncertainty profiles of human ground truth against model-generated continuations on two corpora of professionally-authored short fiction. Using the same LLM as both generator and evaluator of the uncertainty of human text, we compute token- and sequence-level uncertainty metrics to directly quantify the distributional gap between human and LLM storytelling.

\noindent\textbf{RQ1: Do LLMs generate stories with less uncertainty than their human ground truth?} Yes. Across all models, datasets, and metrics, we observe a consistent human–model uncertainty gap. Current post-training alignment methods widen this disparity: instruct and thinking models exhibit significantly larger gaps than their base counterparts.

\noindent\textbf{RQ2: Is this gap unique to creative writing?} It is present across other domains but amplified in creative writing, especially on context-dependent uncertainty metrics.

\noindent\textbf{RQ3: Is uncertainty correlated with writing quality?} Yes. Higher uncertainty correlates consistently with higher quality scores; quadratic regression further reveals diminishing returns at the extremes, consistent with a possible ``sweet spot'' of optimal writing quality at high but not maximal entropy levels.

\section{Uncertainty and Creative Writing}
\label{sec:theory}
Humanities scholarship has historically viewed literature as a pragmatic framework for sense-making via the negotiation of uncertainty \citep{burke1941,iser1978act,cave2016,moi2017}. This section delineates literary-theoretical frameworks that formulate uncertainty as the epistemic locus of meaning-making in literature.

\paragraph{Indeterminacy.}
Literary works possess inevitable ``spots of indeterminacy'' (\textit{Unbestimmtheitsstellen}) where the text remains partially undefined, requiring the reader to actively concretize the aesthetic object \citep{ingarden1973literary}. This incompleteness is not a deficit to be resolved by better alignment with factuality, but the very condition that allows a work to simulate life rather than merely describe it. In the incompleteness of literary language, meaning is constantly ``deferred'' (\textit{différance}), as the ``free play'' of signification ensures that the text resists grounding in a single, monolithic ground truth \citep{derrida1967writing}. Literary insight often resides in the \textit{aporia} (an irresolvable internal contradiction), where effective interpretation requires an epistemic leap into conceptual innovations not yet verifiable by currently available information \citep{deman1979allegories,hartman1980criticism}. Consequently, the ``hallucination'' or semantic drift penalized in post-training is structurally analogous to this ``infinite play'' \citep{barthes1970sz} required for high-fidelity creative expression. By enforcing convergence toward a single, coherent output, LLMs risk collapsing this dynamic instability.

\paragraph{Ambiguity.}
While indeterminacy concerns the inner workings of the text, ambiguity describes the tension between conflicting meanings: if a poetic message can be neatly summarized with minimal loss (i.e., compressed into a ``helpful'' abstract), it ceases to be poetry \citep{brooks1947well}. Thus, the ``fundamental division'' of opposing values in a text attains special aesthetic significance, compelling the reader to hold contradictory concepts in a dialectical suspension—a cognitive state likened to a ``gridiron'' of meaning \citep{empson1930ambiguity}. Literary meaning can thus be formalized as structures of multiplicity, where parallel possibilities coexist to push readers beyond their own experience to foster ethical engagement \citep{serpell2014seven}. Whereas current NLP paradigms typically penalize ambiguity as a failure mode of instruction-following that impedes effective alignment \citep{li2025ambiguity,wang2025learning,li2025condambigqa}, literary theory frames it as a generative ``blessing'' that entangles the reader in the active, interminable construction of meaning \citep{bauer2023strategies}.

\paragraph{Co-creativity.}
The functional utility of uncertainty lies in its capacity to activate the reader's active participation in meaning-making. Leveraging the ``gaps'' created by indeterminacy and ambiguity, a text compels the reader's cognitive engagement to bridge meanings through an interplay of memory and anticipation \citep{iser1978act}. A work lacking such gaps fails as an aesthetic object, making it a ``readerly'' product that renders the audience a passive consumer; conversely, a ``writerly'' text preserves plurality and invites the reader to become a co-creator \citep{barthes1970sz}. Aesthetic openness has also been directly linked to information theory: an ``open work'' is characterized by an increase in information content and entropy, and the refusal of a single, univocal path of interpretation \citep{eco1962open}. LLMs aligned with typicality bias would effectively function as engines of ``readerly'' texts, which risk collapsing openness as a co-creative workspace and resulting in what literary scholars call ``stuplimity'': a synthesis of shock and boredom from the proliferation of frictionless but creatively flattened data \citep{ngai2005ugly}. In this ``textpocalypse'' \citep{Kirschenbaum2023}, the sheer volume of ``post-artificial'' content exhausts the reader not through difficulty, but the smoothness of homogeneous style that invalidates notions of authorship \citep{bajohr2024artificial}.

\paragraph{Failure and Error.}
Recent theory reframes failure not as a performance gap but a counter-hegemonic aesthetic. Through the concept of ``unbecoming,'' \citet{halberstam2011queer} argues that the willingness to fail allows for the unraveling of rigid scripts, challenging success metrics defined by accumulation and mastery. Against the grain of technofuturism's promise to remove friction, ``glitch poetics'' posits errors as revelations of the underlying structures of a digital system, creating a critical form of sensory realism that disrupts the intended seamlessness of the interface \citep{jones2022glitch}. Similarly, the ``poetics of error'' in disability studies highlights how linguistic errors—stuttering, mishearing, breakage, and other qualities typically smoothed over in fined-tuned models \citep{hutchinson2020social}—could generate novel aesthetic possibilities that normative fluency suppresses \citep{davidson2022distressing}. The optimization for instruction-following may inadvertently scrub the generative friction required for these divergent modes of creativity, as the ``errors'' LLMs are trained to avoid could also represent a potential escape from the homogenizing consensus of the training data.

\section{Experiments}
To empirically investigate the role of uncertainty in creative writing, we conduct a controlled story completion study comparing the uncertainty of human-written versus model-generated sequences under identical settings. Beyond text generation, we leverage LLMs as probabilistic measurement tools to estimate the uncertainty of human-authored ground truth. This framework enables the direct quantification of the distributional distinctiveness of human story continuations relative to their machine-generated counterparts.

\subsection{Problem Formulation}

Let $T = T_1 \oplus T_2$ denote a complete text sequence (e.g., a story) with context prefix $T_1$ and continuation $T_2$. 
Given a fixed $T_1$, we compare the uncertainty profiles of two completions:
\begin{enumerate}
    \item \textbf{Human continuation}: the ground-truth $T_2$ as originally written by the human author. Following \citet{yang2023face} and \citet{liu2025evaluating}, we compute the uncertainty of human-written text by evaluating the likelihood of the original text's sequence $T_2$ under the model $P_\theta$ by recording the model's predictive distribution over each token position in a provided prompt of the full story $T = T_1 \oplus T_2$. Specifically, for each token $t_i \in T_2$, we extract the token-level log-probability given context $p_i = P_\theta(t_i \mid T_1, t_{<i})$. This measures the ``surprisal'' or unpredictability of human creativity from the model's perspective.
    \item \textbf{Model continuation}: a generated sequence $\hat{T}_2 \sim P_\theta(\cdot \mid T_1)$ sampled from an LLM. For all sampled tokens $t_i \in \hat{T}_2$, we similarly record the token-level log-probability given context $p_i$ at each generation step. 
\end{enumerate}

\paragraph{Uncertainty Metrics.} 
We quantify uncertainty using both token-level surprisal metrics such as Negative Log-Likelihood (NLL) and Perplexity (PPL), alongside sequence-level measures of contextual dependence like Pointwise Mutual Information (PMI) \citep{church1990word} and Conditional PMI \citep{vanderpoel2022mutual}. Because the marginal subtraction in PMI and CPMI makes them more robust against potential self-scoring bias and surface-level lexical rarity, we treat them as our primary evidence and only report NLL and PPL as descriptive measures of raw surprisal. A comprehensive mathematical formulation and detailed interpretation of these metrics are provided in Appendix~\ref{app:uncertainty}.

Crucially, we use the \emph{same} LLM $P_\theta$ as both the generator and the evaluator of the uncertainty of human text. This ensures that any observed differences in uncertainty arise from the texts $T_2$ and $\hat{T}_2$ themselves rather than from the distributional mismatch between evaluation and generation models. To further ensure a fair comparison, we constrain the generation length such that $\hat{T}_2$ has exactly $|T_2|$ tokens, matching the length of the human continuation. While we expect this design might lead the model to underreport raw surprisal (NLL, PPL) on its own outputs (since $\hat{T}_2$ is sampled from $P_\theta$ while $T_2$ is not), we address this potential confound by including context-normalized metrics—PMI and CPMI cancel out any self-scoring advantage as they subtract the marginal $\log P_\theta(t_i \mid t_{<i})$ on a per-token basis. Our design is validated empirically by experiments with a held-out evaluator in Appendix~\ref{app:heldout}.

This direct comparison setup allows us to answer the question: \emph{From the perspective of an LLM, how surprising is the human-written story compared to its own continuation?} Following the intuitions of literary theory, we hypothesize a \textbf{human--model uncertainty gap} $\Delta_{\mathcal{H}}$: since LLMs (especially instruct models) are trained to be less comfortable with ambiguity, they are likely to exhibit lower intrinsic uncertainty and complete the story in a less uncertain way than the human ground truth. Several studies in computational creativity have described human-model differences as a gap \citep{moon2025homogenizing,xu2025echoes,peeperkorn2025mind}; to the best of our knowledge, we are the first to quantify such gaps in creative writing via information-theoretic approaches.

\subsection{Datasets}
\label{sec:datasets}
We conduct our analysis on two high-fidelity short story datasets (Table~\ref{tab:dataset_stats}):

\begin{itemize}
    \item \textbf{The New Yorker}: we curate all short stories published from 2010 to 2019 in \textit{The New Yorker}, widely considered one of the most prestigious venues for literary fiction.
    \item \textbf{Tell-Me-A-Story} \citep{huot2025agents}: a high-quality corpus of human-authored short stories spanning diverse genres and narrative styles. Unlike typical web-scraped creative writing data, these stories were collected through targeted writing workshops where skilled writers collaborated on complex, open-ended prompts.
\end{itemize}

By limiting our analysis to professional writers and excluding crowd-sourced lay platforms (e.g., r/WritingPrompts), we ensure data quality while minimizing the risk of training data memorization. Additionally, we filter stories exceeding 4096 tokens to accommodate model context windows and prevent truncation of the cumulative $T_1$ (Section~\ref{sec:setup}).

\subsection{Experimental Setup}
\label{sec:setup}

\paragraph{Model Selection.} 
We evaluate our hypothesis across a diverse suite of open-weights models as the underlying probability distribution $P_\theta$, spanning the \textsc{OLMo}, \textsc{Llama}, \textsc{Qwen}, \textsc{Mistral}, \textsc{Gemma}, and \textsc{Phi} families. To isolate the impact of post-training alignment on creative uncertainty, we pair each base model with its instruct-tuned counterpart. We extend this comparison to include reasoning-enhanced ``thinking'' variants when an exact base-instruct-thinking triplet is available (e.g., \textsc{OLMo-3}, \textsc{Qwen-3}).

\paragraph{Sequence Segmentation.} 
To prevent context fragmentation from confounding our uncertainty estimates, we treat $T_1$ as a cumulative prefix rather than a truncated sliding window. The latter approach risks discarding critical early context, which would unfairly bias the comparison between model and human uncertainty. Specifically, for a ground truth story with sentences $(s_1, s_2, \ldots, s_m)$, we construct context--continuation pairs $(T_1^{(k)}, T_2^{(k)})$ where $T_1^{(k)} = s_1 \oplus \cdots \oplus s_{k-1}$ and $T_2^{(k)} = s_k$. This setup yields multiple data points per document with progressively longer contexts, ensuring that the model retains the full narrative history.

\paragraph{Implementation Details.} 
We utilize \texttt{vLLM} \citep{kwon2023efficient} for high-throughput inference. All experiments are conducted in \texttt{bfloat16} precision using standard sampling: we set $\texttt{temperature}=1.0$ to ensure the stochasticity required for a fair comparison with human writing, and disable nucleus sampling
\footnote{While nucleus sampling ($p = 0.95$ or $0.9$) is standard practice to maintain coherence, we intentionally disable it to establish a conservative baseline. Because tail truncation artificially reduces output entropy, utilizing nucleus sampling would only artificially widen the observed human-model uncertainty gap.} 
($p = 1.0$) to guarantee that the output faithfully reflects the model's true distribution without artificial tail truncation. Due to constraints in the \texttt{vLLM} v1 engine, accurate log-probabilities are available only for the top-20 tokens. For ground-truth tokens falling outside this set, we impute a lower-bound estimate using a dynamic floor of $\min(\text{observed logprobs}) - 2.0$ nats.

\paragraph{Evaluation.}
For each completion pair $(T_2, \hat{T}_2)$, we compute the ratio ($\text{Human} / \text{Model}$) for NLL and PPL to measure relative surprisal, and the difference  ($\text{Human} - \text{Model}$) for PMI and CPMI to capture deviations in contextual determination. This paired setup normalizes for inter-model differences, enabling a direct comparison of human versus model uncertainty with the same underlying distribution.

\section{Results}
\label{sec:results}

We present the results of our comparative analysis in Table~\ref{tab:main_results}. Across all datasets, uncertainty metrics,  and the full range of model families, scales, and post-training variants, we observe a pervasive \textbf{human--model uncertainty gap}: when given identical contexts, LLMs consistently generate story continuations $\hat{T}_2$ with substantially lower intrinsic uncertainty than the human-authored ground truth $T_2$. The uniformly negative $\Delta$PMI values ($-1.44$ to $-2.60$) shows that $\hat{T}_2$ exhibits stronger context-conditioned predictability—models over-rely on predefined narrative context to select statistically probable next tokens, whereas human authors introduce higher surprisal by deviating from expected narrative paths. $\Delta$CPMI shows more inter-model variance but aligns directionally with other metrics, indicating that at high-entropy decision points, human choices remain less predictable than model outputs. This trend is further corroborated by token-level metrics: the median NLL ratio ranges from 2.03 to 3.9, indicating that human-written stories are on average 2–4$\times$ more ``surprising'' to the model than its own generations; perplexity ratios amplify this gap to 2.76–8.82$\times$.

The uncertainty gap is notably wider for the \textit{New Yorker} dataset than \textit{Tell-Me-A-Story} (Table~\ref{tab:main_results}). The \textit{New Yorker} is widely regarded as a gold standard of contemporary literary fiction, often featuring experimental narratives with high ambiguity and semantic economy. In terms of literary production, these stories likely represent a later stage of editorial refinement compared to the \textit{Tell-Me-A-Story} corpus, where writers drafted stories from scratch in a workshop. This discrepancy suggests a correlation between writing quality and uncertainty: professionally curated literary fiction maintains higher novelty and structural complexity, creating a distribution that is distinct from workshop-level writing and significantly harder for aligned LLMs to approximate.

\begin{table*}[t]
\caption{(Top) New Yorker, (Bottom) Tell-Me-A-Story. Values reported are medians. NLL and PPL are reported as ratios (human/model); PMI and CPMI as differences (human$-$model), negative $\Delta$PMI indicates higher human uncertainty. Cell colors represent human–model gap magnitude (green = small, red = large). Instruct and thinking models exhibit significantly larger gaps than their base variants.}
\label{tab:main_results}
\centering
\small
\resizebox{\textwidth}{!}{%
\begin{tabular}{lcccc|cccc|cccc}
\toprule
 & \multicolumn{4}{c|}{\textbf{Base}} & \multicolumn{4}{c|}{\textbf{Instruct}} & \multicolumn{4}{c}{\textbf{Thinking}} \\
\cmidrule(r){2-5} \cmidrule(lr){6-9} \cmidrule(l){10-13}
\textbf{Model} & \textbf{NLL} & \textbf{PPL} & \textbf{PMI} & \textbf{CPMI} & \textbf{NLL} & \textbf{PPL} & \textbf{PMI} & \textbf{CPMI} & \textbf{NLL} & \textbf{PPL} & \textbf{PMI} & \textbf{CPMI} \\
\midrule
OLMo-2-7B & \cellcolor{gap2}2.18 & \cellcolor{gap3}3.62 & \cellcolor{gap2}-1.70 & \cellcolor{gap1}0.13 & \cellcolor{gap5}3.01 & \cellcolor{gap4}5.97 & \cellcolor{gap3}-1.89 & \cellcolor{gap2}0.01 &  &  &  &  \\
OLMo-2-13B & \cellcolor{gap2}2.19 & \cellcolor{gap2}3.48 & \cellcolor{gap2}-1.63 & \cellcolor{gap1}0.12 & \cellcolor{gap3}2.48 & \cellcolor{gap3}4.14 & \cellcolor{gap3}-1.77 & \cellcolor{gap1}0.10 &  &  &  &  \\
OLMo-2-32B & \cellcolor{gap2}2.15 & \cellcolor{gap2}3.19 & \cellcolor{gap1}-1.55 & \cellcolor{gap1}0.08 & \cellcolor{gap3}2.32 & \cellcolor{gap3}3.51 & \cellcolor{gap2}-1.64 & \cellcolor{gap1}0.07 &  &  &  &  \\
OLMo-3-7B & \cellcolor{gap2}2.20 & \cellcolor{gap3}3.95 & \cellcolor{gap3}-1.81 & \cellcolor{gap1}0.15 & \cellcolor{gap5}3.30 & \cellcolor{gap5}7.98 & \cellcolor{gap5}-2.20 & \cellcolor{gap1}0.09 & \cellcolor{gap4}2.74 & \cellcolor{gap5}6.07 & \cellcolor{gap5}-2.17 & \cellcolor{gap1}0.11 \\
OLMo-3-32B & \cellcolor{gap2}2.15 & \cellcolor{gap2}3.47 & \cellcolor{gap2}-1.66 & \cellcolor{gap1}0.09 & \cellcolor{gap4}2.99 & \cellcolor{gap4}5.59 & \cellcolor{gap2}-1.68 & \cellcolor{gap3}-0.43 & \cellcolor{gap4}2.78 & \cellcolor{gap4}5.06 & \cellcolor{gap2}-1.66 & \cellcolor{gap3}-0.40 \\
\addlinespace
Llama-3.1-8B & \cellcolor{gap2}2.17 & \cellcolor{gap3}3.55 & \cellcolor{gap5}-2.49 & \cellcolor{gap3}-0.42 & \cellcolor{gap2}2.13 & \cellcolor{gap3}3.76 & \cellcolor{gap5}-2.60 & \cellcolor{gap3}-0.25 &  &  &  &  \\
\addlinespace
Qwen3-4B & \cellcolor{gap3}2.21 & \cellcolor{gap3}4.07 & \cellcolor{gap3}-1.72 & \cellcolor{gap1}0.14 & \cellcolor{gap5}3.28 & \cellcolor{gap5}7.37 & \cellcolor{gap2}-1.56 & \cellcolor{gap1}0.29 & \cellcolor{gap5}3.32 & \cellcolor{gap5}7.84 & \cellcolor{gap1}-1.55 & \cellcolor{gap1}0.24 \\
\addlinespace
Phi-4 &  &  &  &  & \cellcolor{gap3}2.27 & \cellcolor{gap3}3.84 & \cellcolor{gap2}-1.67 & \cellcolor{gap1}0.13 & \cellcolor{gap3}2.37 & \cellcolor{gap3}4.32 & \cellcolor{gap2}-1.65 & \cellcolor{gap1}0.11 \\
Phi-4 (Reasoning+) &  &  &  &  &  &  &  &  & \cellcolor{gap3}2.42 & \cellcolor{gap3}4.44 & \cellcolor{gap2}-1.60 & \cellcolor{gap1}0.10 \\
\addlinespace
Mistral-Small-24B & \cellcolor{gap2}2.12 & \cellcolor{gap2}3.21 & \cellcolor{gap4}-2.03 & \cellcolor{gap3}-0.48 & \cellcolor{gap1}2.07 & \cellcolor{gap2}3.16 & \cellcolor{gap4}-2.03 & \cellcolor{gap4}-0.57 &  &  &  &  \\
\addlinespace
Gemma-3-4B & \cellcolor{gap3}2.21 & \cellcolor{gap3}3.90 & \cellcolor{gap4}-2.06 & \cellcolor{gap4}-0.58 & \cellcolor{gap5}3.90 & \cellcolor{gap5}8.82 & \cellcolor{gap4}-1.95 & \cellcolor{gap5}-1.06 &  &  &  &  \\
Gemma-3-12B & \cellcolor{gap2}2.15 & \cellcolor{gap2}3.43 & \cellcolor{gap4}-1.91 & \cellcolor{gap4}-0.62 & \cellcolor{gap5}3.29 & \cellcolor{gap5}6.18 & \cellcolor{gap3}-1.87 & \cellcolor{gap5}-0.98 &  &  &  &  \\
Gemma-3-27B & \cellcolor{gap3}2.33 & \cellcolor{gap3}3.58 & \cellcolor{gap3}-1.78 & \cellcolor{gap4}-0.79 & \cellcolor{gap5}3.60 & \cellcolor{gap5}6.50 & \cellcolor{gap4}-1.94 & \cellcolor{gap4}-0.85 &  &  &  &  \\
\midrule
\midrule
\midrule
OLMo-2-7B & \cellcolor{gap2}2.15 & \cellcolor{gap2}3.15 & \cellcolor{gap1}-1.55 & \cellcolor{gap2}0.04 & \cellcolor{gap5}3.22 & \cellcolor{gap4}5.48 & \cellcolor{gap3}-1.85 & \cellcolor{gap2}0.01 &  &  &  &  \\
OLMo-2-13B & \cellcolor{gap2}2.16 & \cellcolor{gap2}3.05 & \cellcolor{gap1}-1.48 & \cellcolor{gap2}0.03 & \cellcolor{gap4}2.59 & \cellcolor{gap3}3.83 & \cellcolor{gap2}-1.69 & \cellcolor{gap1}0.05 &  &  &  &  \\
OLMo-2-32B & \cellcolor{gap2}2.13 & \cellcolor{gap1}2.91 & \cellcolor{gap1}-1.45 & \cellcolor{gap2}0.01 & \cellcolor{gap3}2.42 & \cellcolor{gap2}3.42 & \cellcolor{gap2}-1.56 & \cellcolor{gap2}0.04 &  &  &  &  \\
OLMo-3-7B & \cellcolor{gap2}2.14 & \cellcolor{gap2}3.35 & \cellcolor{gap2}-1.66 & \cellcolor{gap1}0.05 & \cellcolor{gap5}3.51 & \cellcolor{gap5}7.25 & \cellcolor{gap5}-2.12 & \cellcolor{gap1}0.08 & \cellcolor{gap4}2.75 & \cellcolor{gap4}5.14 & \cellcolor{gap4}-2.01 & \cellcolor{gap2}0.04 \\
OLMo-3-32B & \cellcolor{gap2}2.11 & \cellcolor{gap1}2.98 & \cellcolor{gap1}-1.50 & \cellcolor{gap2}0.00 & \cellcolor{gap5}3.03 & \cellcolor{gap4}4.82 & \cellcolor{gap2}-1.59 & \cellcolor{gap3}-0.45 & \cellcolor{gap4}2.81 & \cellcolor{gap3}4.42 & \cellcolor{gap1}-1.55 & \cellcolor{gap3}-0.42 \\
\addlinespace
Llama-3.1-8B & \cellcolor{gap1}2.09 & \cellcolor{gap2}3.01 & \cellcolor{gap5}-2.35 & \cellcolor{gap3}-0.50 & \cellcolor{gap1}2.09 & \cellcolor{gap2}3.23 & \cellcolor{gap5}-2.48 & \cellcolor{gap3}-0.34 &  &  &  &  \\
\addlinespace
Qwen3-4B & \cellcolor{gap2}2.16 & \cellcolor{gap2}3.47 & \cellcolor{gap1}-1.55 & \cellcolor{gap2}0.03 & \cellcolor{gap5}3.37 & \cellcolor{gap5}6.41 & \cellcolor{gap1}-1.48 & \cellcolor{gap1}0.23 & \cellcolor{gap5}3.37 & \cellcolor{gap5}6.68 & \cellcolor{gap1}-1.44 & \cellcolor{gap1}0.19 \\
\addlinespace
Phi-4 &  &  &  &  & \cellcolor{gap3}2.23 & \cellcolor{gap2}3.30 & \cellcolor{gap1}-1.52 & \cellcolor{gap2}0.03 & \cellcolor{gap3}2.33 & \cellcolor{gap3}3.70 & \cellcolor{gap2}-1.57 & \cellcolor{gap2}0.00 \\
Phi-4-Reasoning+ &  &  &  &  &  &  &  &  & \cellcolor{gap3}2.36 & \cellcolor{gap3}3.73 & \cellcolor{gap1}-1.50 & \cellcolor{gap2}0.00 \\
\addlinespace
Mistral-Small-24B & \cellcolor{gap1}2.07 & \cellcolor{gap1}2.79 & \cellcolor{gap4}-1.91 & \cellcolor{gap3}-0.54 & \cellcolor{gap1}2.03 & \cellcolor{gap1}2.76 & \cellcolor{gap4}-1.94 & \cellcolor{gap4}-0.63 &  &  &  &  \\
\addlinespace
Gemma-3-4B & \cellcolor{gap2}2.13 & \cellcolor{gap2}3.31 & \cellcolor{gap4}-1.97 & \cellcolor{gap4}-0.68 & \cellcolor{gap5}3.81 & \cellcolor{gap5}6.95 & \cellcolor{gap4}-1.91 & \cellcolor{gap5}-1.12 &  &  &  &  \\
Gemma-3-12B & \cellcolor{gap1}2.07 & \cellcolor{gap1}2.97 & \cellcolor{gap3}-1.83 & \cellcolor{gap4}-0.71 & \cellcolor{gap5}3.20 & \cellcolor{gap4}5.03 & \cellcolor{gap3}-1.86 & \cellcolor{gap5}-1.01 &  &  &  &  \\
Gemma-3-27B & \cellcolor{gap3}2.25 & \cellcolor{gap2}3.10 & \cellcolor{gap3}-1.71 & \cellcolor{gap4}-0.82 & \cellcolor{gap5}3.57 & \cellcolor{gap4}5.38 & \cellcolor{gap4}-1.92 & \cellcolor{gap4}-0.84 &  &  &  &  \\
\bottomrule
\end{tabular}
}
\end{table*}

Comparing model architectures, \textsc{Mistral-Small-24B} has the overall smallest human--model gap, while \textsc{Gemma-3-27B} (Base) and \textsc{Gemma-3-4B} (Instruct) show the largest
\footnote{Interestingly, \textsc{Gemma-3} models suffer from particularly high uncertainty gaps, implying a uniquely rigid generation distribution—a pattern consistent with the wide use of knowledge distillation in the \textsc{Gemma} family \citep{team2024gemma}, which is observed to  reduce the student model's intrinsic entropy and illicit mode-seeking behavior \citep{gu2024minillm,cha2025why}}
. While NLL and PPL rankings are highly consistent across all models, they sometimes directionally disagree with PMI, revealing tradeoffs between local and global uncertainty. For instance, \textsc{Llama-3.1-8B-Instruct} records the largest $\Delta$PMI ($-2.6$ on \textit{New Yorker} / $-2.48$ on \textit{Tell-Me-A-Story}), indicating high context over-reliance, despite achieving the best token-level alignment (NLL ratio $2.13$ / $2.09$). Conversely, \textsc{Qwen3-4B-Instruct} displays the smallest $\Delta$PMI ($-1.56$), suggesting its generations are less bounded by contextual expectations, even though it ranks among the highest in NLL and PPL ratio as it fails to match human-level local uncertainty. This divergence highlights that surprisal-based and context-dependence metrics capture complementary aspects of the human--model distributional mismatch. Appendix~\ref{app:pmi_cpmi} quantifies this relationship by decomposing $\Delta$CPMI into a surprisal component and a decision-point context-dependence component, whose relative size determines the sign of $\Delta$CPMI.

Within model families, instruction-tuned and ``thinking'' variants
\footnote{We manually verify that all instruct/reasoning outputs are valid story continuations and required no additional parsing.}
consistently exacerbate the uncertainty gap compared to base models, which produce the most human-like uncertainty profiles. Paired Wilcoxon tests across stories confirm this widening for 18--20 of 22 family$\times$dataset cells on NLL/PPL with $|r_{rb}| \approx 1$, and for all 6 base-to-thinking cells (Appendix~\ref{app:significance}). The magnitude of this alignment tax further varies across families: \textsc{Mistral-Small-24B}'s instruct variant achieves near-parity with the base model, and \textsc{Llama-3.1-8B} demonstrates relative stability (NLL ratios of $2.17$ vs. $2.13$); these two are also the only families with no significant NLL/PPL widening in our paired tests. In contrast, \textsc{Gemma-3-4B}'s NLL and PPL ratios double from base to instruct, and various models from the \textsc{OLMo} family also suffer from a substantial 1.3–1.5$\times$ ``inflation''. With the exception of \textsc{OLMo-3-7B}, reasoning-enhanced thinking models provide little recovery and occasionally widen the gap to base variants, and additional reasoning compute (e.g., \textsc{Phi-4-Reasoning} vs. \textsc{Plus}) has minimal impact on creative uncertainty.
Model scale has little effect on the base-model uncertainty gap, where NLL ratios change by at most $0.13$ between the extreme sizes within a family, but it partially mitigates the alignment tax that post-training imposes—within \textsc{OLMo-2}, the base-instruct gap (NLL) shrinks from 38\% at \textsc{7B} to 8\% at \textsc{32B}, and the same monotone shrinkage holds for \textsc{OLMo-3} and \textsc{Gemma-3} (Appendix~\ref{app:scale}). Our observed differences between intra-family variants align with recent studies showing that base models preserve greater creative diversity than instruct variants \citep{west2025base}, and chain-of-thought reasoning primarily benefits mathematical and symbolic tasks rather than open-ended generation \citep{sprague2025cot}.

\section{Discussion}
\subsection{Is the Uncertainty Gap Domain-Specific?}
\label{sec:domain}
To investigate whether the observed human--model uncertainty gap is unique in creative writing or generalizes across other domains, we replicate our experimental setup on the Ghostbuster dataset \citep{verma2024ghostbuster}, which provides a controlled contrast between human-authored narrative texts from three distinct genres (creative writing, news, and essays; details see Table~\ref{tab:dataset_stats}). We randomly sample documents from each domain (matched for corpus size) and apply the identical settings of our main experiment
\footnote{Due to limited compute, we confine the scope of this experiment to instruct models under 14B.}
. Comparing these domains allows us to test the uncertainty gap is amplified in creative contexts where ambiguity and indeterminacy are intrinsic features, as opposed to factual or structured narratives in functional domains where the ``correct'' continuation is more constrained.

\begin{table}[t]
\caption{Uncertainty gap by domain. Values are medians aggregated across 8 instruct models on the Ghostbuster corpus. The domain gap is very pronounced on $\Delta$PMI and $\Delta$CPMI.}
\label{tab:domain_comparison}
\centering
\small
\begin{tabular}{lcccc}
\toprule
\textbf{Domain} & \textbf{NLL} & \textbf{PPL} & \textbf{PMI ($\Delta$)} & \textbf{CPMI ($\Delta$)} \\
\midrule
Essays & 3.07 & 6.03 & $-1.71$ & $-0.09$ \\
News (Reuters) & 3.31 & 5.91 & $-1.73$ & $-0.10$ \\
Creative Writing & 3.11 & 6.60 & $\mathbf{-2.15}$ & $\mathbf{-0.28}$ \\
\bottomrule
\end{tabular}
\end{table}

Table~\ref{tab:domain_comparison} summarizes the human--model uncertainty gap across domains. While the gap is present in all three domains, it is frequently exacerbated in uncertainty metrics that focus on open-ended creative contexts. Notably, the $\Delta$PMI gap is 25--30\% wider for creative writing ($-2.15$) compared to essays ($-1.71$) and news ($-1.73$). Since PMI measures how strongly the context $T_1$ predicts the continuation $T_2$, this pattern suggests that models exhibit significantly greater context over-reliance when generating continuations for stories compared to functional narrative prose. While models struggle to match human uncertainty broadly, they are particularly constrained in deviating from contextual expectations in open-ended settings where such deviations constitute a core feature of literary expression.

In contrast, token-level metrics are relatively stable across domains, with NLL ratios ranging from 2.9--3.3$\times$ and PPL between 5.9--6.6$\times$, a token-level gap that seems to be a more general property of aligned LLMs. This dissociation between local and global uncertainty patterns indicates that the creative-specific amplification of the gap is not driven by surface-level lexical choices but rather by deeper structural phenomena: models generate locally predictable tokens at comparable rates across domains, yet they fail to introduce the narrative-level tension and surprisal that characterize high-quality human storytelling. The $\Delta$CPMI gap gap further corroborates this finding, showing the largest magnitude ($-0.28$) for creative writing, where high-entropy decision points (moments of genuine narrative ambiguity) are disproportionately resolved by models toward safe, generic outcomes. Full per-model results are provided in Appendix~\ref{app:domain_results}.

\subsection{Does Uncertainty Correlate with Writing Quality?}
\label{sec:quality}

One potential objection to our findings is that the high uncertainty observed in human writing may merely represent stochastic noise or a lack of clarity, rather than a desirable aesthetic quality. To distinguish between arbitrary randomness and the constructive ambiguity and indeterminacy required for literary richness, we empirically examine whether the uncertainty gap translates into a qualitative difference in writing performance.

We leverage the \textsc{Writing Quality Reward Model} \citep{chakrabarty2025aislop}, a ModernBERT-based discriminator fine-tuned on expert preferences of creative writing pairs, to assign scalar quality scores to both human ground truth $(T_1, T_2)$ and model continuations $(T_1, \hat{T}_2)$.
\footnote{Inputs are obtained by concatenating each context–continuation pair, filtering for passages between 150--400 words to align with the length of the reward model's training data.}
Using these quality estimates as the dependent variable, we first test for monotonic association via Spearman correlation, then probe for non-linear structure via quadratic regression ($y = \beta_0 + \beta_1 x + \beta_2 x^2$) against our uncertainty metrics (NLL, PPL, and PMI). We conduct these analyses across all models and datasets, evaluating human-authored and model-generated continuations separately. Given that human creativity often follows the Wundt curve where aesthetic value peaks at a moderately high level of novelty before degrading into incoherence \citep{berlyne1971aesthetics}, the quadratic term specifically probes whether there exists a ``sweet spot'' of optimal uncertainty.

\begin{table}[t]
\centering
\small
\caption{Relationship between uncertainty metrics and writing quality. $\rho$: mean Spearman correlation $\pm$ std; Sig.: \% of models with significant correlation in the expected direction (positive for NLL/PPL, negative for PMI; $p < 0.05$); Sweet Spot: \% exhibiting a significant inverted-U (quadratic) relationship with peak in range.}
\label{tab:quality_correlations}
\begin{tabular}{llccccc}
\toprule
\textbf{Source} & \textbf{Metric} & $\rho$ & \textbf{Sig.} & \textbf{\% Linear} & \textbf{Sweet Spot} & \textbf{\% Diminishing} \\
\midrule
\multirow{3}{*}{Human} 
  & NLL & $+0.072 \pm 0.049$ & 56\% & 12\% & 34\% & 9\% \\
  & PPL & $+0.072 \pm 0.049$ & 56\% & 5\%  & 0\%  & 25\% \\
  & PMI & $-0.106 \pm 0.052$ & 75\% & 44\% & 2\%  & 0\% \\
\midrule
\multirow{3}{*}{Model} 
  & NLL & $+0.109 \pm 0.062$ & 77\% & 28\% & 31\% & 19\% \\
  & PPL & $+0.109 \pm 0.062$ & 77\% & 9\%  & 31\% & 33\% \\
  & PMI & $-0.076 \pm 0.040$ & 69\% & 45\% & 2\%  & 0\% \\
\bottomrule
\end{tabular}
\end{table}

\paragraph{Uncertainty--quality relationship.}
Table~\ref{tab:quality_correlations} summarizes the results across all models and datasets. The correlations are small in absolute terms ($|\bar{\rho}| = 0.07$--$0.11$), but highly consistent across 28 models and 3 datasets. The Spearman correlations reveal a near-uniform monotonic positive association between token-level uncertainty (NLL, PPL) and quality scores: for human-authored continuations, the mean $\bar{\rho}=+0.072$ ($\sigma=0.049$), with 56\% of model--dataset combinations yielding significant positive correlations ($p<0.05$). Model-generated continuations exhibit a similar but slightly stronger pattern ($\bar{\rho}=+0.109$, 77\% significant). In contrast, PMI shows a robust negative correlation with quality for both human ($\bar{\rho}=-0.106$, 75\% significantly negative) and model text ($\bar{\rho}=-0.076$, 69\% significantly negative). Recall that a lower PMI indicates a weaker dependence on the immediate context $T_1$. This result suggests that contextual divergence is consistently associated with quality: the more a continuation creates its own information distinct from the strong priors of the prompt, the higher it is rated. Models with high PMI (context over-reliance) indicate strong contextual coherence, which may correspond to more formulaic or predictable prose that resembles auto-completing the most probable clich\'{e} implied by the prompt, whereas high-quality writing necessitates a departure from these obvious trajectories.

The shape classification in Table~\ref{tab:quality_correlations} refines this monotonic picture by revealing non-linearity in a substantial subset of combinations. For human-authored NLL, 34\% of model--dataset combinations exhibit a statistically significant inverted-U (``sweet spot''), with quality peaking at $\bar{z}^*=1.98$ SDs above the mean uncertainty—indicating that the optimum lies at high but not extreme levels of unpredictability. Model continuations show a comparable pattern (31\% sweet-spot for NLL, peak at $z^*=1.48$). This non-monotonic relationship is most pronounced in \textit{Tell-Me-A-Story}, where 79\% of models show sweet-spot behavior for human text (see Appendix~\ref{app:quality_detailed} for per-dataset breakdowns). However, the additional variance explained by the quadratic term is small ($\overline{\Delta R^2} < 0.01$), indicating that the dominant pattern is a positive monotonic trend with gentle curvature rather than a sharp quality reversal at high uncertainty. The rightward-shifted peaks suggest that the diminishing-returns region lies near the high-entropy tail, consistent with the view that skilled human writing operates near the edge of unpredictability without crossing into incoherence.

Crucially, the vast majority of instruct models in our study operate to the left of these peaks. By minimizing the KL-divergence from safe, modal outputs during alignment, these models are effectively constrained to a low-entropy regime that mechanistically precludes the uncertainty required for high-quality narrative. While some human writing also demonstrates this positive correlation and shows a logarithmic curve of diminishing returns rather than a full drop-off, human writers rarely produce the low-entropy, repetitive text characteristic of over-aligned models.

\paragraph{Writers vs.\ LLMs.}
The uncertainty--quality relationship differs systematically between human and model-generated text. Human continuations show weaker but more consistent correlations across datasets, with sweet-spot peaks occurring at higher uncertainty levels ($z^* = 1.98$) compared to model text ($z^* = 1.48$). This gap suggests that human writers naturally operate in a higher-entropy regime where quality gains from additional unpredictability are more modest, while model outputs (clustered in a lower-entropy band) show steeper quality gradients as they escape their modal tendencies. The per-dataset breakdown (Table~\ref{tab:quality_detailed}) reveals that this pattern is most pronounced in \textit{New Yorker}, where human correlations are near zero ($\rho = 0.021$) but model correlations remain moderate ($\rho = 0.053$), in line with the hypothesis that professionally edited human prose already occupies the quality-optimal uncertainty range. Notably, while model continuations show stronger correlations with quality than human text ($\Delta\rho=0.037$ for NLL), this does not imply that higher uncertainty uniformly benefits LLM-generated stories. Rather, the pattern likely reflects the restricted variance of model outputs: model continuations cluster in a narrower uncertainty range, and within this range, those samples that do exhibit higher entropy tend to escape the ``slop'' of generic, high-probability text. The full per-model results (Table~\ref{tab:quality_per_model}) and shape distribution analysis (Table~\ref{tab:shape_distribution}) are provided in Appendix~\ref{app:quality_detailed}.

\section{Conclusion}

This paper provides the first systematic, literary- and information-theoretic formalization of the role of uncertainty in creative writing. We demonstrate that LLMs consistently generate story continuations with 2--4$\times$ lower entropy and substantially higher context-dependence than human-authored ground truth—a gap that widens under post-training alignment and persists across model families and scales. Crucially, the uncertainty gap is amplified in creative writing relative to functional domains, and strongly correlates with writing quality.

Our results highlight the need for uncertainty-aware alignment strategies that can mitigate the destructive ambiguity of a hallucinated medical diagnosis while preserving the constructive ambiguity of a poetic metaphor, and distinguish between the two. To achieve parity with human writers, future work must explore how to re-align computational systems to tolerate, and indeed generate, the specific forms of pragmatic uncertainties outlined in literary theory: possible interventions include enriching pre-training mixtures with curated fiction, training SFT on diverse self-continuations, developing domain-conditional reward models that distinguish destructive from constructive ambiguity, and calibrating entropy-aware decoding to the quality sweet spots ($z^* \approx 1.5$--$2.0$) our regressions identify. As a potential guide for these interventions, our framework offers a reproducible, stage-level diagnostic that transforms the subjective intuition of ``flat'' LLM prose into a measurable benchmark for future alignment research. A full discussion of causal mechanisms, stage-specific recommendations, and limitations is provided in Appendix~\ref{app:supplemental_discussion}.

\section*{Ethics Statement}
This work involves no human subjects and utilizes publicly available or archived literary datasets. To ensure compliance, the \textit{New Yorker} data is stored in a secure repository at our university; its use in this study constitutes fair use, as it is employed strictly for non-disseminating, non-commercial research. Other corpora, such as Tell-Me-a-Story, were sourced from open-source projects intended for research use.

\paragraph{Humanistic Mission Statement.} 
This paper seeks to advance the intersection of NLP and the digital humanities. As LLMs integrate deeper into society, NLP research are becoming increasingly interdisciplinary, encountering nuanced problem spaces that prove recalcitrant to the expertise and methods of computer science alone. We argue that bridging NLP research with existing domain knowledge in relevant disciplines (e.g., literary theory for computational creativity) is vital. Just as the field has long embraced physics-informed machine learning for grounding models in the real world, we advocate for the development of literary-theory-informed models for computational creativity. We hope to encourage more collaborative work between NLP researchers and humanities scholars, ensuring that future models are grounded in the rigorous theoretical frameworks of the domains they attempt to emulate.

\bibliography{colm2026_conference}
\bibliographystyle{colm2026_conference}

\newpage
\appendix

\section{Uncertainty Estimation Methods}
\label{app:uncertainty}

We quantify the uncertainty of $T_2$ and $\hat{T}_2$ with several token and sequence-level metrics (Table~\ref{tab:metrics}) derived from the model's log-probabilities. 

\begin{table}[b]
\caption{Summary of uncertainty metrics and their interpretation.}
\label{tab:metrics}
\centering
\small
\begin{tabular}{lcc}
\toprule
\textbf{UE Metric} & \textbf{Higher Value Indicates} & \textbf{Scope} \\
\midrule
Token Entropy & More uncertainty & Token \\
Perplexity & More uncertainty & Sequence \\
PMI & More predictable by context & Sequence \\
CPMI & Uncertainty at key positions & Sequence \\
\bottomrule
\end{tabular}
\end{table}

\paragraph{Mean Token Entropy (Surprisal).}
We approximate the entropy of a text sequence using length-normalized Negative Log-Likelihood (NLL), which serves as a proxy for the average information content per token. For a sequence with tokens $(t_1, \ldots, t_n)$ given context $T_1$, we compute the mean surprisal as:

$$
\mathcal{H}_{\text{token}} = -\frac{1}{n} \sum_{i=1}^{n} \log P_\theta(t_i \mid T_1, t_{<i})
$$

This quantity, established in information-theoretic linguistics \citep{hale2001probabilistic, levy2008expectation}, reflects the model's average ``unexpectedness'' of each token in the story continuation.

\paragraph{Perplexity.} 
We compute the sequence perplexity as the exponentiated mean token surprisal, as a normalized measure of the model's predictive uncertainty:
$$
\text{PPL} = \exp\left(\mathcal{H}_{\text{token}}\right)
$$

\paragraph{Pointwise Mutual Information (PMI).} 
To disentangle context-dependent uncertainty from the effects of intrinsic token frequency (i.e., high-probability words), we compute the Pointwise Mutual Information between the context $T_1$ and the continuation $T_2$. PMI normalizes the conditional likelihood by an unconditional baseline:

$$
\text{PMI}(T_1; T_2) = \frac{1}{n}\sum_{i=1}^{n} \left[ \log P_\theta(t_i \mid T_1, t_{<i}) - \log P_\theta(t_i \mid t_{<i}) \right]
$$

where $t_{<i}$ denotes the prefix within the continuation $T_2$. The second term represents the unconditional log-probability, estimated via a secondary inference pass over $T_2$ without the context $T_1$. For the model continuation $\hat{T}_2$, the unconditional marginals $P_\theta(t_i \mid t_{<i})$ are similarly computed with a second pass over the generated sequence $\hat{T}_2$ itself.

This metric distinguishes between uncertainty from creative content versus generic linguistic complexity from model priors (e.g., difficult syntax). A high PMI indicates that tokens in $T_2$ are specifically relevant to the story context $T_1$; conversely, a near-zero or negative value suggests the context does not provide significant predictive signal for the continuation, potentially indicating a disassociation from the prompt or lateral shift in the narrative.

\paragraph{Conditional PMI (CPMI).} 
Standard PMI can be biased by high-frequency stopwords. Following \citet{vanderpoel2022mutual}, we additionally compute a thresholded variant that focuses on high-entropy tokens:
$$
\text{CPMI} = \mathcal{H}_{\text{token}} + \frac{1}{n} \sum_{i: \mathcal{H}_i \geq \tau} \log P_\theta(t_i \mid t_{<i})
$$
where $\mathcal{H}_i = -\log P_\theta(t_i \mid T_1, t_{<i})$ is the per-token surprisal, and we use $\lambda = 1.0$, $\tau = 2.0$ nats throughout. 
Our adoption of CPMI utilizes a threshold-based formulation where the unconditional term is only summed if the conditional entropy exceeds a specific calibration threshold $\tau$ (experimentally set to $\tau=2.0$ nats). This metric upweights the contribution of unconditional probabilities for tokens that are already uncertain given context.

\subsection{Top-K Log-Probability Imputation}
\label{app:topk}

The \texttt{vLLM} v1 engine returns exact log-probabilities only for the top-20 tokens at each position (Section~\ref{sec:setup}). When a ground-truth token falls outside that set, we impute a floor of $\min(\text{observed logprobs}) - 2.0$ nats, i.e.\ the 20th-ranked log-probability minus 2 nats. Human continuations contain more out-of-top-20 tokens than model continuations, so this approximation deserves scrutiny. Its effect, however, runs opposite to the direction one might expect, and leaves the reported gap conservative.

The concern would be that the imputed value sits \emph{below} the true log-probability, penalizing those tokens too harshly and inflating human NLL. In practice the imputed value sits \emph{above} the truth for the deep-tail tokens that dominate the out-of-top-20 case in natural text. The rank distribution past the top of the softmax is heavy-tailed, and a token at rank 1{,}000 typically carries a log-probability many nats below the 20th-ranked token. The imputation therefore over-states the probability of these tokens and under-states their contribution to NLL. Because human continuations contain more of them, the reported human NLL is under-stated more aggressively than the model NLL, which makes the reported gap a lower bound on the true gap.

PMI and CPMI are less sensitive still to the choice of $K$. Both subtract $\log P_\theta(t_i \mid t_{<i})$ from $\log P_\theta(t_i \mid T_1, t_{<i})$ on a per-token basis. When a token falls outside the top-20 under both conditional and unconditional scoring, the common case for rare lexical items, the same imputed value enters both terms and cancels in the difference.

Memorization points the same way. It is unlikely in our setting, since the \textit{New Yorker} stories are proprietary and \textit{Tell-Me-A-Story} was released in 2025, but it cannot be fully ruled out, and a model that saw a human story during pre-training will assign it higher probability and lower NLL. Methods that reliably disentangle memorization from uncertainty estimation would yield more accurate human--model gaps, which our analysis suggests would be larger than those Table~\ref{tab:main_results} reports.

\section{Decomposition of the CPMI Gap}
\label{app:pmi_cpmi}

$\Delta$PMI is uniformly negative in Table~\ref{tab:main_results}, while $\Delta$CPMI changes sign across families. The two metrics measure the same construct, the human--model gap in context-dependence, but over different parts of the sequence. $\Delta$PMI averages the gap across every token in the continuation, whereas CPMI \citep{vanderpoel2022mutual} adds the unconditional term only at tokens whose conditional surprisal exceeds $\tau = 2$ nats; everywhere else CPMI behaves like the conditional surprisal NLL. When the two disagree, we are observing which conditional-surprisal stratum carries the gap: high-entropy decision points, or low-entropy filler.

This can be made exact. Writing $S := \text{CPMI} - \text{NLL}$ for the decision-point piece of CPMI, the identity
$$
\Delta\text{CPMI} \;=\; \Delta\text{NLL}_{\text{abs}} + \Delta S
$$
holds at the row level, where $\Delta\text{NLL}_{\text{abs}}$ is the absolute (rather than ratio) surprisal gap. We verified it numerically across all 56 cells of Table~\ref{tab:main_results}, with a maximum residual of $7 \times 10^{-15}$. The two components read cleanly. $\Delta\text{NLL}_{\text{abs}}$ is positive throughout our data, meaning humans are more surprising than models. $\Delta S$ is the same context-dependence gap that $\Delta$PMI measures but restricted to decision-point tokens, and it is negative throughout, meaning humans rely on context more at uncertain moments. The sign of $\Delta$CPMI follows from which term is larger in magnitude, and the ratio $|\Delta S| / |\Delta\text{PMI}|$ localizes the gap: below 1 the gap is spread across all tokens including low-entropy filler, near 1 it is uniform across token types, and above 1 it is concentrated at decision points.

Table~\ref{tab:pmi_cpmi_decomp} reports the decomposition for every cell. This appendix uses the same window-level aggregation as Table~\ref{tab:main_results}, so the recomputed NLL ratio, $\Delta$PMI, and $\Delta$CPMI columns reproduce the reported medians to two decimal places. Note that the per-row identity is what holds exactly; at the median level it is only approximate, since the median of a sum is not the sum of medians.

\begin{table*}[t]
\centering
\footnotesize
\caption{Decomposition of the CPMI gap for every cell of Table~\ref{tab:main_results}. $\Delta\text{NLL}_{\text{abs}}$ is the absolute surprisal gap (human $-$ model) and $\Delta S$ is the decision-point component, which sums unconditional log-probabilities only at tokens whose conditional surprisal exceeds $\tau = 2$ nats. Per row, $\Delta\text{CPMI} = \Delta\text{NLL}_{\text{abs}} + \Delta S$. The ratio $|\Delta S|/|\Delta\text{PMI}|$ localizes the gap, and the localization tag summarizes it: filler below ${\approx}0.9$, uniform near 1, decision points above ${\approx}1.05$. The tag agrees across both datasets for all 28 cells.}
\label{tab:pmi_cpmi_decomp}
\resizebox{\textwidth}{!}{%
\begin{tabular}{ll|ccccc|ccccc|l}
\toprule
& & \multicolumn{5}{c|}{\textbf{New Yorker}} & \multicolumn{5}{c|}{\textbf{Tell-Me-A-Story}} & \\
\cmidrule(lr){3-7} \cmidrule(lr){8-12}
\textbf{Family} & \textbf{Variant} & $\Delta$PMI & $\Delta$CPMI & $\Delta\text{NLL}_{\text{abs}}$ & $\Delta S$ & $\frac{|\Delta S|}{|\Delta \text{PMI}|}$ & $\Delta$PMI & $\Delta$CPMI & $\Delta\text{NLL}_{\text{abs}}$ & $\Delta S$ & $\frac{|\Delta S|}{|\Delta \text{PMI}|}$ & \textbf{Localization} \\
\midrule
OLMo-2-7B & Base & $-$1.70 & +0.13 & +1.29 & $-$1.14 & 0.67 & $-$1.55 & +0.04 & +1.15 & $-$1.08 & 0.70 & filler \\
OLMo-2-7B & Instruct & $-$1.89 & +0.01 & +1.79 & $-$1.79 & 0.95 & $-$1.85 & +0.01 & +1.70 & $-$1.70 & 0.92 & uniform \\
\addlinespace
OLMo-2-13B & Base & $-$1.63 & +0.12 & +1.25 & $-$1.11 & 0.68 & $-$1.48 & +0.03 & +1.11 & $-$1.07 & 0.72 & filler \\
OLMo-2-13B & Instruct & $-$1.77 & +0.10 & +1.42 & $-$1.30 & 0.74 & $-$1.69 & +0.05 & +1.34 & $-$1.27 & 0.75 & filler \\
\addlinespace
OLMo-2-32B & Base & $-$1.55 & +0.08 & +1.16 & $-$1.07 & 0.69 & $-$1.45 & +0.01 & +1.07 & $-$1.04 & 0.72 & filler \\
OLMo-2-32B & Instruct & $-$1.64 & +0.07 & +1.26 & $-$1.18 & 0.72 & $-$1.56 & +0.04 & +1.23 & $-$1.18 & 0.76 & filler \\
\addlinespace
OLMo-3-7B & Base & $-$1.81 & +0.15 & +1.37 & $-$1.21 & 0.67 & $-$1.66 & +0.05 & +1.21 & $-$1.15 & 0.69 & filler \\
OLMo-3-7B & Instruct & $-$2.20 & +0.09 & +2.08 & $-$1.99 & 0.91 & $-$2.12 & +0.08 & +1.98 & $-$1.89 & 0.89 & uniform \\
OLMo-3-7B & Thinking & $-$2.17 & +0.11 & +1.80 & $-$1.69 & 0.78 & $-$2.01 & +0.04 & +1.64 & $-$1.61 & 0.80 & filler \\
\addlinespace
OLMo-3-32B & Base & $-$1.66 & +0.09 & +1.24 & $-$1.14 & 0.69 & $-$1.50 & +0.00 & +1.09 & $-$1.08 & 0.72 & filler \\
OLMo-3-32B & Instruct & $-$1.68 & $-$0.43 & +1.72 & $-$2.20 & 1.31 & $-$1.59 & $-$0.45 & +1.57 & $-$2.10 & 1.33 & decision pts \\
OLMo-3-32B & Thinking & $-$1.66 & $-$0.40 & +1.62 & $-$2.08 & 1.25 & $-$1.55 & $-$0.42 & +1.49 & $-$1.98 & 1.28 & decision pts \\
\addlinespace
Llama-3.1-8B & Base & $-$2.49 & $-$0.42 & +1.27 & $-$1.67 & 0.67 & $-$2.35 & $-$0.50 & +1.10 & $-$1.62 & 0.69 & filler \\
Llama-3.1-8B & Instruct & $-$2.60 & $-$0.25 & +1.33 & $-$1.59 & 0.61 & $-$2.48 & $-$0.34 & +1.17 & $-$1.52 & 0.61 & filler \\
\addlinespace
Qwen3-4B & Base & $-$1.72 & +0.14 & +1.40 & $-$1.25 & 0.73 & $-$1.55 & +0.03 & +1.25 & $-$1.20 & 0.77 & filler \\
Qwen3-4B & Instruct & $-$1.56 & +0.29 & +2.00 & $-$1.70 & 1.09 & $-$1.48 & +0.23 & +1.86 & $-$1.61 & 1.09 & decision pts \\
Qwen3-4B & Thinking & $-$1.55 & +0.24 & +2.06 & $-$1.78 & 1.15 & $-$1.44 & +0.19 & +1.90 & $-$1.69 & 1.17 & decision pts \\
\addlinespace
Phi-4 & Instruct & $-$1.67 & +0.13 & +1.35 & $-$1.21 & 0.72 & $-$1.52 & +0.03 & +1.19 & $-$1.16 & 0.76 & filler \\
Phi-4 & Thinking & $-$1.65 & +0.11 & +1.46 & $-$1.35 & 0.82 & $-$1.57 & +0.00 & +1.31 & $-$1.30 & 0.83 & filler \\
\addlinespace
Phi-4 (Reasoning+) & Thinking & $-$1.60 & +0.10 & +1.49 & $-$1.38 & 0.86 & $-$1.50 & +0.00 & +1.32 & $-$1.31 & 0.87 & filler \\
\addlinespace
Mistral-Small-24B & Base & $-$2.03 & $-$0.48 & +1.17 & $-$1.65 & 0.82 & $-$1.91 & $-$0.54 & +1.03 & $-$1.58 & 0.83 & filler \\
Mistral-Small-24B & Instruct & $-$2.03 & $-$0.57 & +1.15 & $-$1.74 & 0.86 & $-$1.94 & $-$0.63 & +1.02 & $-$1.67 & 0.86 & filler \\
\addlinespace
Gemma-3-4B & Base & $-$2.06 & $-$0.58 & +1.36 & $-$1.99 & 0.97 & $-$1.97 & $-$0.68 & +1.20 & $-$1.92 & 0.98 & uniform \\
Gemma-3-4B & Instruct & $-$1.95 & $-$1.06 & +2.18 & $-$3.33 & 1.71 & $-$1.91 & $-$1.12 & +1.94 & $-$3.15 & 1.65 & decision pts \\
\addlinespace
Gemma-3-12B & Base & $-$1.91 & $-$0.62 & +1.23 & $-$1.90 & 0.99 & $-$1.83 & $-$0.71 & +1.09 & $-$1.83 & 1.00 & uniform \\
Gemma-3-12B & Instruct & $-$1.87 & $-$0.98 & +1.82 & $-$2.86 & 1.53 & $-$1.86 & $-$1.01 & +1.62 & $-$2.69 & 1.45 & decision pts \\
\addlinespace
Gemma-3-27B & Base & $-$1.78 & $-$0.79 & +1.27 & $-$2.11 & 1.19 & $-$1.71 & $-$0.82 & +1.13 & $-$2.01 & 1.18 & decision pts \\
Gemma-3-27B & Instruct & $-$1.94 & $-$0.85 & +1.87 & $-$2.79 & 1.44 & $-$1.92 & $-$0.84 & +1.68 & $-$2.59 & 1.35 & decision pts \\
\bottomrule
\end{tabular}
}
\end{table*}

Three regimes emerge, illustrated below with New Yorker values (Tell-Me-A-Story shows the same pattern). Where the decision-point gap dominates, $\Delta$CPMI is strongly negative: \textsc{Gemma-3-4B} Instruct has $|\Delta S|/|\Delta\text{PMI}| = 1.71$, and $\Delta S = -3.33$ swamps $\Delta\text{NLL}_{\text{abs}} = +2.18$ to give $\Delta\text{CPMI} = -1.06$. Where the gap is roughly uniform, the two components cancel: \textsc{OLMo-2-7B} Instruct has a ratio of ${\approx}0.95$ with $\Delta\text{NLL}_{\text{abs}} = +1.79$ against $\Delta S = -1.79$, so $\Delta\text{CPMI} \approx 0$. Where the average gap exceeds the decision-point gap, $\Delta$CPMI is mildly negative: \textsc{Llama-3.1-8B} Base has a ratio of $0.67$, and $\Delta S = -1.67$ narrowly beats $\Delta\text{NLL}_{\text{abs}} = +1.27$ for $\Delta\text{CPMI} = -0.42$. This accounts for the pattern in Table~\ref{tab:main_results}: $\Delta$CPMI is positive for the families whose surprisal gap dominates (base \textsc{OLMo-2} and \textsc{OLMo-3}, \textsc{Qwen3}, \textsc{Phi-4}) and negative for those whose decision-point gap dominates (\textsc{Llama}, \textsc{Mistral}, every \textsc{Gemma} size, post-trained \textsc{OLMo-3-32B}).

The two metrics are therefore complementary rather than contradictory. $\Delta$PMI summarizes context-dependence on average; $\Delta$CPMI summarizes its net effect at narrative decision points, after accounting for the model's surprisal disadvantage. Because our sequences are stories, the decomposition also indicates which parts of a text contribute most to the token-level context-dependence gap. Whether these strata map onto narrative structure, for instance junctures where many continuations are plausible, would require qualitative analysis that we leave to future work.

\section{Held-out Evaluator Validation}
\label{app:heldout}

Because each model both generates and scores its own continuations (Section~\ref{sec:setup}), $\hat{T}_2$ is drawn from $P_\theta$ while $T_2$ is not, and part of the raw surprisal gap reflects the evaluator's familiarity with its own output. To separate that component from a property of the texts, we re-scored the continuations of four instruct models under an independent evaluator.

\paragraph{Setup.} We re-score the continuations of \textsc{Qwen3-4B-Instruct-2507}, \textsc{OLMo-3-7B-Instruct}, \textsc{Gemma-3-4B-it}, and \textsc{Gemma-3-12B-it}, together with their corresponding human continuations, under \textsc{Llama-3.1-8B-Instruct}. All context--continuation pairs from both datasets are covered: 9,074 for \textit{New Yorker} and 22,992 for \textit{Tell-Me-A-Story} per source model. Apart from the evaluator, the procedure matches the main experiment: \texttt{vLLM} in \texttt{bfloat16} with \texttt{max\_model\_len}=4096 and \texttt{prompt\_logprobs}=20; conditional scoring tokenizes $T_1 \oplus T_2$ and reads the $T_2$ span, unconditional scoring tokenizes $T_2$ alone; CPMI uses $\lambda = 1.0$ and $\tau = 2.0$ nats. The evaluator is itself one of the models in Table~\ref{tab:main_results}, but it is not among the four source models re-scored here, so no row in this experiment is self-scored.

\begin{table}[t]
\centering
\small
\caption{Medians under an independent \textsc{Llama-3.1-8B-Instruct} evaluator. Unlike Table~\ref{tab:main_results}, values are raw per-source medians rather than human/model ratios or differences; the gap is the human row minus (or over) each model row. Human NLL and PPL exceed every model on both datasets, and human PMI/CPMI fall below every model, matching the direction of Table~\ref{tab:main_results}.}
\label{tab:heldout}
\begin{tabular}{lcccc|cccc}
\toprule
& \multicolumn{4}{c|}{\textbf{New Yorker}} & \multicolumn{4}{c}{\textbf{Tell-Me-A-Story}} \\
\cmidrule(lr){2-5} \cmidrule(lr){6-9}
\textbf{Source} & \textbf{NLL} & \textbf{PPL} & \textbf{PMI} & \textbf{CPMI} & \textbf{NLL} & \textbf{PPL} & \textbf{PMI} & \textbf{CPMI} \\
\midrule
Human (ground truth) & 3.11 & 22.53 & 1.31 & $-$0.24 & 2.78 & 16.08 & 1.52 & $-$0.21 \\
\midrule
\textsc{Qwen3-4B-Instruct} & 2.85 & 17.29 & 1.73 & 0.09 & 2.56 & 12.95 & 2.05 & 0.25 \\
\textsc{OLMo-3-7B-Instruct} & 3.06 & 21.28 & 1.77 & 0.08 & 2.71 & 15.03 & 2.12 & 0.24 \\
\textsc{Gemma-3-4B-it} & 2.74 & 15.42 & 1.82 & 0.19 & 2.39 & 10.88 & 2.25 & 0.37 \\
\textsc{Gemma-3-12B-it} & 2.74 & 15.50 & 1.93 & 0.21 & 2.41 & 11.16 & 2.35 & 0.41 \\
\bottomrule
\end{tabular}
\end{table}

\paragraph{Results.} The gap is directionally validated, though 3 of 4 metrics shrink substantially in magnitude (Table~\ref{tab:heldout}).

\begin{itemize}
    \item \textbf{NLL and PPL gaps shrink the most.} The 2--4$\times$ human/model NLL ratio under self-scoring (Table~\ref{tab:main_results}) now ranges only between 1.02 and 1.16, and PPL between 1.06 and 1.48. Human NLL and PPL nonetheless remain higher than every model on both datasets.
    \item \textbf{PMI gaps are better preserved and more consistent across families.} The metric's construction explains this: because PMI subtracts the marginal log-probability on a per-token basis, it does not lean on the evaluator's familiarity with its own outputs the way NLL does. The held-out run thus empirically verifies the evaluator-robustness property we relied on in Section~\ref{sec:setup} and Appendix~\ref{app:uncertainty} when adopting PMI as a metric.
    \item \textbf{CPMI gaps become \emph{more} consistent under held-out scoring.} In Table~\ref{tab:main_results} the self-scored $\Delta\text{CPMI} = \Delta\text{NLL}_{\text{abs}} + \Delta S$ varies in sign and magnitude across families (Appendix~\ref{app:pmi_cpmi}). Under the held-out evaluator $\Delta\text{NLL}_{\text{abs}}$ collapses, and the residual decision-point component $\Delta S$ produces uniformly negative $\Delta$CPMI gaps.
\end{itemize}

A substantial share of the self-scored NLL/PPL advantage therefore reflects the model's familiarity with its own outputs rather than a property of the continuations. What survives an independent scorer is carried by the context-normalized metrics, which is why Sections~\ref{sec:setup} and~\ref{sec:results} present PMI and CPMI as the load-bearing evidence and NLL/PPL as descriptive.

\paragraph{Why the single-model setup remains primary.} With this validation in place, we keep the design of Table~\ref{tab:main_results} because it offers concrete advantages for three of our research aims that a held-out evaluator would obscure. First, it gives a cleaner base versus post-trained comparison: each variant scoring its own continuations isolates what post-training did to that model's own distribution, whereas a single fixed evaluator would contaminate the comparison with its own training artifacts, and a panel of cross-referenced models would complicate it unnecessarily. Second, it gives a more direct cross-corpus comparison of human texts: when contrasting human corpora, as in the cross-domain comparison with Ghostbuster, keeping each model as its own evaluator isolates how unfamiliar each corpus is to that specific model, while a held-out evaluator could mix its own corpus-specific idiosyncrasies into the contrast. Third, it keeps the relationship with writing-quality metrics (Section~\ref{sec:quality}) interpretable, since cross-model scoring would make the correlations and regressions considerably harder to read. The single-model setup is thus the cleanest read on within-model and within-corpus comparisons, and the held-out run shows that the PMI/CPMI signal it produces is preserved under independent scoring.

\section{Supplemental Discussion}
\label{app:supplemental_discussion}
This paper provides the first systematic, literary- and information-theoretic formalization of the role of uncertainty in creative writing. We demonstrate that LLMs consistently generate story continuations with substantially higher context-dependence than human-authored ground truth, and with 2--4$\times$ lower entropy under self-scoring. The gap widens under post-training alignment, persists across model families and scales, and survives re-scoring by an independent evaluator on the context-normalized metrics. It is also amplified in creative writing relative to functional domains, and shows a consistent though modest correlation with writing quality. We now discuss what these findings imply about \emph{why} the gap arises, \emph{where} interventions are most needed, and \emph{how} our quantitative framework can guide them.

\paragraph{Why does the uncertainty gap arise?}
Our results point to post-training alignment as the primary amplifier rather than sole cause. The gap already exists for base models (NLL ratios of ${\sim}2.1\text{--}2.3\times$), suggesting that pre-training on predominantly expository, deduplicated web text (where ambiguity is incidental rather than functional) already biases models toward low-entropy distributions. However, the near-doubling of NLL and PPL ratios after instruction tuning (e.g., \textsc{Gemma-3-4B}: $2.21 \rightarrow 3.90$) implicates the standard post-training pipeline as the dominant contributor. Two specific mechanisms are consistent with our evidence. First, the KL-divergence penalty in RLHF explicitly penalizes divergence from a reference policy, compressing the output distribution toward modal, ``safe'' completions—precisely the low-entropy regime our quality analysis identifies as suboptimal. Second, reward models trained on general-purpose human preferences systematically favor verbalized confidence and unambiguous helpfulness \citep{leng2025taming,zhang2025verbalized}, properties antithetical to the constructive ambiguity that our literary-theoretic framework identifies as necessary for narrative richness. The near-parity of base and instruct
variants for \textsc{Mistral-Small-24B} and \textsc{Llama-3.1-8B} demonstrates that this ``alignment tax'' is not inevitable—it is a design choice amenable to targeted intervention.

\paragraph{Where should we intervene?}
Our findings suggest a stage-differentiated strategy. At \emph{pre-training}, the base-model gap (${\sim}2\times$) indicates that current data curation pipelines (which prioritize deduplication, factual consistency, and domain balance) may systematically under-represent the high-entropy, ambiguity-rich register of literary prose. Enriching pre-training mixtures with editorially curated fiction corpora could narrow this baseline gap, though care must be taken to avoid memorization. At \emph{supervised fine-tuning (SFT)}, our observation that PMI gaps widen more than token-level gaps after alignment points to the instruction-following objective as a driver of context over-reliance rather than mere lexical narrowing: SFT data for creative tasks could incorporate diverse, non-obvious continuations rather than single ``gold'' references, explicitly training models to sustain narrative-level surprisal. At \emph{RLHF}, the most impactful lever may be \emph{domain-conditional reward modeling}: our evidence that the gap is significantly larger in creative writing than in news or essays ($\Delta\text{PMI}$: $-2.15$ vs.\ $-1.71$) motivates reward functions that can distinguish between destructive ambiguity (e.g., a hallucinated medical dosage) and constructive ambiguity (e.g., an unresolved narrative tension), penalizing the former while preserving or encouraging the latter. Finally, at \emph{inference time}, our quality--uncertainty regressions identify sweet-spot peaks at $z^* \approx 1.5$--$2.0$ standard deviations above mean entropy, offering a concrete, empirically grounded target for entropy-aware decoding strategies (e.g., dynamically adjusting sampling temperature or minimum-entropy thresholds conditioned on the detected domain and narrative context).

\paragraph{What does quantification of this gap enable?}
While the intuition that LLM-generated stories feel ``flat'' is widely shared, intuition alone cannot diagnose the problem, measure progress, or compare interventions. Our information-theoretic metrics transform this subjective impression into a reproducible diagnostic instrument. The uncertainty gap provides (i) an \emph{automatic, reference-based benchmark} for evaluating whether a new alignment method, dataset, or decoding strategy brings model outputs closer to the distributional profile of professional human writing; (ii) a \emph{stage-level diagnostic} that can attribute distributional collapse to specific training phases by comparing base, SFT, and RLHF checkpoints of the same model; and (iii) a \emph{domain-sensitive probe} that captures the amplified challenge of creative generation beyond what surface-level diversity metrics (e.g., distinct-$n$, Self-BLEU) detect. The consistent correlation between uncertainty and writing quality (Section~\ref{sec:quality}) further indicates that these metrics track a property that matters for downstream literary value, not merely statistical divergence.

\paragraph{Relation to machine-generated text detection.}
The signals we measure sit adjacent to those that recent detection methods exploit: DetectGPT \citep{mitchell2023detectgpt} reads the local curvature of model log-probabilities, and Binoculars \citep{hans2024binoculars} thresholds a between-model perplexity ratio. Detection is not our focus, and our own results argue against treating the gap on its own as a sufficient detection signal, for two reasons. The gap is domain-dependent: it is largest on creative writing and shrinks substantially on more conventional text (Section~\ref{sec:domain}), so a gap-driven detector would be best calibrated for creative-domain detection rather than general-purpose identification, which is still useful given the volume of AI-generated literary content. The gap is also adversarially fragile: surprisal-based signatures are susceptible to decoding-time intervention, and temperature tuning, nucleus sampling, and contrastive decoding that deliberately broadens the output distribution could close the gap without changing what the model is doing, while light paraphrasing would perturb token-level surprisal in ways that may defeat thresholding. A robust detector built on this signal would need to combine the distributional gap with adversarially aware methods and domain-specific calibration, neither of which we address here.

\paragraph{Limitations and future work.}
Our framework quantifies the gap but does not yet close it. Validating the stage-specific interventions outlined above requires controlled ablations across training pipelines, which we leave to future work. Our reliance on a single reward model for quality estimation, while a practical design choice to avoid evaluation confounds, should be complemented by expert human judgments in future studies. Additionally, our analysis focuses on English-language short fiction; extending the framework to other literary traditions, multilingual settings, and longer-form narratives (e.g., novels) would test the generality of our claims. We also acknowledge that uncertainty is a necessary but likely not sufficient condition for literary quality—other dimensions such as narrative coherence, character development, and thematic depth interact with uncertainty in ways our current metrics do not fully capture.

Four further limitations follow from the analyses in this appendix. Token-level metrics alone cannot fully separate \emph{constructive} ambiguity from simple low model likelihood; formalizing what ``constructive'' means in a creative context is a question for computational pragmatics that will require methods beyond uncertainty estimation. Plot-level uncertainty, the controlled release of information that drives suspense and recognition, operates at the event and span level, a register token-level metrics cannot speak to directly; \citet{wilmot2020modelling} model suspense at that register, and we see information-theoretic measures of the kind we propose as a substrate that cognitive accounts can build on rather than a substitute for them. The cross-domain comparison in Section~\ref{sec:domain} rests on a single corpus and may depend on its genre, style, and source distribution, so we hope to expand it into a full-scale study covering a wider variety of domain-specific datasets. Finally, our qualitative reading of what high-uncertainty human text contains (Appendix~\ref{app:methodological_notes}) has not been validated by a formal human evaluation, which remains a keen interest for future work.

Achieving parity with human writers requires that future alignment research move beyond the binary of hallucination-as-failure toward a more nuanced taxonomy of uncertainty that recognizes its functional role in creative expression. Our framework takes a first step by providing the measurement infrastructure to guide this effort: a quantitative foundation for uncertainty-aware alignment paradigms that can preserve the constructive ambiguity necessary for literary richness while mitigating the destructive ambiguity that undermines factual reliability.

\section{Datasets}
\label{app:datasets}

\begin{table}[h]
\centering
\caption{Dataset Statistics}
\label{tab:dataset_stats}
\renewcommand{\arraystretch}{3}
\resizebox{\columnwidth}{!}{%
\begin{tabular}{lcccccccc}
\toprule
\textbf{Dataset} & \makecell{\textbf{Total}\\\textbf{Stories}} & \makecell{\textbf{After}\\\textbf{Filtering}} & \textbf{Sentences} & \textbf{Tokens} & \textbf{Writers} & \textbf{Data Source} & \textbf{Availability} & \makecell{\textbf{Memorization}\\\textbf{Risk}} \\
\midrule
New Yorker & 504 & 70 & 9,144 & 201,199 & \makecell{Expert, some\\prize-winning} & \makecell{\textit{The New Yorker}\\magazine} & Private & Very Low \\
\makecell[l]{Tell\_Me\_a\_Story\\\citep{huot2025agents}} & 230 & 226 & 23,218 & 500,452 & Professional & Story workshop & \makecell{Public (requires\\decryption)} & Low \\
\makecell[l]{Ghostbuster\\\citep{verma2024ghostbuster}} & 299 & 298 & 10,010 & 213,749 & Lay & \makecell{Reddit\\(r/WritingPrompts)} & Public & Medium \\
\bottomrule
\end{tabular}%
}
\end{table}

Table~\ref{tab:dataset_stats} summarizes the statistics of all datasets used in our experiments. Below we provide additional details on data collection, curation, and filtering procedures.

\subsection{Primary Datasets}

\paragraph{The New Yorker.}
We curated all short stories published in \textit{The New Yorker} between 2010 and 2019. \textit{The New Yorker} is widely regarded as one of the most prestigious venues for literary fiction in the English-speaking world, with a rigorous editorial selection process that accepts only a small fraction of submissions. Many stories in this collection have received major literary awards or been anthologized in ``Best American Short Stories'' compilations. Due to copyright restrictions, we do not release this dataset publicly. The low memorization risk stems from robust copyright protection and the fact that these stories are behind a paywall, making them unlikely to appear in large-scale web crawls used for LLM pretraining.

\paragraph{Tell-Me-A-Story.}
The \textit{Tell-Me-A-Story} corpus \citep{huot2025agents} is the only publicly-available story dataset sourced from professional writer that we could find. It represents a departure from typical web-scraped creative writing data. Stories were collected through structured writing workshops where skilled writers, including MFA students, published authors, and writing instructors; they collaborated to produce narratives based on complex, open-ended prompts designed to elicit diverse genres and narrative styles. The workshop setting encouraged writers to take creative risks and produce polished work, resulting in higher average quality compared to crowd-sourced alternatives. While the dataset is publicly available, it requires decryption and access approval, reducing the likelihood of inclusion in pretraining corpora. 

\subsection{Domain Generalization Dataset}

\paragraph{Ghostbuster.}
To investigate whether the human--model uncertainty gap generalizes beyond literary fiction into functional and information-seeking domains, we utilize the Ghostbuster dataset \citep{verma2024ghostbuster}, which provides human-authored texts across three distinct genres:

\begin{itemize}
    \item \textbf{Creative Writing:} Short fiction sourced from Reddit's r/WritingPrompts community. While these writers are not professionally trained, the subreddit's voting mechanism surfaces higher-quality submissions. We use this subset to compare against our primary datasets.
    \item \textbf{News Articles:} Journalism samples representing factual, objective prose with conventional discourse structures.
    \item \textbf{Student Essays:} Academic writing samples with argumentative and expository styles.
\end{itemize}

\section{Is the Uncertainty Gap Domain-Specific?}
\label{app:domain_results}

Table~\ref{tab:domain_full} presents the complete per-model uncertainty metrics across all three domains in the Ghostbuster corpus. Across all models, the $\Delta$PMI gap is consistently larger (more negative) for creative writing than for essays and news articles, while NLL and PPL ratios exhibit less systematic variation by domain.

This asymmetry between the two metric families is what we would expect, rather than evidence against creative-writing specificity. NLL and PPL are readily affected by surface-level features such as lexical rarity, which vary with corpus more than with task. PMI and CPMI instead measure how much a continuation depends on its prefix, and that is the dimension on which functional writing genuinely differs from literary writing: the writer's task in news or student essays is constrained by external facts and rhetorical conventions, leaving less room for context-bound interpretive choices. The widening of $\Delta$PMI from $-1.71$ and $-1.73$ on essays and news to $-2.15$ on creative writing, and of $\Delta$CPMI from roughly $-0.10$ to $-0.28$, is the cross-domain version of the observation that motivated our choice of metrics in the first place. Once the surface-rarity component is stripped away by the marginal subtraction built into PMI and CPMI, the residual context-dependence gap is amplified in creative writing relative to functional domains. The stability of NLL and PPL across domains is therefore the expected baseline against which that amplification reads as the finding of this section.

\begin{table*}[h]
\centering
\small
\caption{Full domain comparison results. All values are medians. NLL and PPL are ratios (human/model); PMI and CPMI are differences (human $-$ model).}
\label{tab:domain_full}
\resizebox{\textwidth}{!}{%
\begin{tabular}{ll|cccc|cccc|cccc}
\toprule
& & \multicolumn{4}{c|}{\textbf{Essays}} & \multicolumn{4}{c|}{\textbf{News (Reuters)}} & \multicolumn{4}{c}{\textbf{Creative Writing}} \\
\textbf{Model} & \textbf{Size} & \textbf{NLL} & \textbf{PPL} & \textbf{PMI} & \textbf{CPMI} & \textbf{NLL} & \textbf{PPL} & \textbf{PMI} & \textbf{CPMI} & \textbf{NLL} & \textbf{PPL} & \textbf{PMI} & \textbf{CPMI} \\
\midrule
OLMo-2-Instruct & 7B & 3.45 & 7.14 & $-1.66$ & 0.24 & 3.46 & 6.16 & $-1.69$ & 0.20 & 3.46 & 7.43 & $-2.13$ & 0.01 \\
OLMo-2-Instruct & 13B & 2.85 & 5.11 & $-1.58$ & 0.25 & 2.86 & 4.48 & $-1.59$ & 0.20 & 2.68 & 4.85 & $-1.95$ & 0.08 \\
OLMo-3-Instruct & 7B & 3.30 & 7.32 & $-1.92$ & 0.25 & 3.62 & 8.07 & $-2.25$ & 0.26 & 3.51 & 8.94 & $-2.33$ & 0.05 \\
\addlinespace
Llama-3.1-Instruct & 8B & 2.15 & 3.79 & $-2.24$ & $-0.19$ & 2.27 & 3.47 & $-1.99$ & $-0.34$ & 2.17 & 3.95 & $-3.03$ & $-0.26$ \\
\addlinespace
Qwen3-Instruct & 4B & 3.25 & 6.60 & $-1.50$ & 0.34 & 3.51 & 6.92 & $-1.47$ & 0.32 & 3.55 & 8.33 & $-1.63$ & 0.21 \\
\addlinespace
Phi-4 & 14B & 2.32 & 3.88 & $-1.55$ & 0.18 & 2.26 & 3.28 & $-1.50$ & 0.08 & 2.28 & 3.94 & $-1.74$ & 0.04 \\
\addlinespace
Gemma-3-it & 4B & 3.96 & 8.20 & $-1.68$ & $-0.90$ & 4.89 & 9.18 & $-1.64$ & $-0.79$ & 3.94 & 8.98 & $-2.21$ & $-1.27$ \\
Gemma-3-it & 12B & 3.30 & 6.17 & $-1.54$ & $-0.85$ & 3.60 & 5.70 & $-1.68$ & $-0.76$ & 3.27 & 6.35 & $-2.17$ & $-1.10$ \\
\midrule
\textbf{Avg. (Median)} & & 3.07 & 6.03 & $-1.71$ & $-0.09$ & 3.31 & 5.91 & $-1.73$ & $-0.10$ & 3.11 & 6.60 & $-2.15$ & $-0.28$ \\
\bottomrule
\end{tabular}
}
\end{table*}

\begin{table}[h]
\centering
\small
\caption{Relative increase in PMI gap for creative writing compared to functional domains.}
\label{tab:pmi_increase}
\begin{tabular}{lcc}
\toprule
\textbf{Model} & \textbf{vs. Essays (\%)} & \textbf{vs. News (\%)} \\
\midrule
OLMo-2-7B-Instruct & +28.3 & +26.0 \\
OLMo-2-13B-Instruct & +23.4 & +22.6 \\
OLMo-3-7B-Instruct & +21.4 & +3.6 \\
Llama-3.1-8B-Instruct & +35.3 & +52.3 \\
Qwen3-4B-Instruct & +8.7 & +10.9 \\
Phi-4 & +12.3 & +16.0 \\
Gemma-3-4B-it & +31.5 & +34.8 \\
Gemma-3-12B-it & +40.9 & +29.2 \\
\midrule
\textbf{Average} & +25.2 & +24.4 \\
\bottomrule
\end{tabular}
\end{table}

\section{Additional Methodological Notes}
\label{app:methodological_notes}

This section provides supplementary details regarding our experimental setup, scope constraints, and additional observations from our results:

\begin{itemize}
    \item \textbf{Nucleus Sampling:} While nucleus sampling ($p = 0.95$ or $0.9$) is standard practice to maintain coherence, we intentionally disable it to establish a conservative baseline. Because tail truncation artificially reduces output entropy, utilizing nucleus sampling would only artificially widen the observed human--model uncertainty gap.
    
    \item \textbf{Output Validity:} For our model continuations, we manually verify that all instruct and reasoning outputs are valid story continuations and required no additional parsing.
    
    \item \textbf{\textsc{Gemma-3} Distribution Rigidity:} Interestingly, \textsc{Gemma-3} models suffer from particularly high uncertainty gaps, implying a uniquely rigid generation distribution---a pattern consistent with the wide use of knowledge distillation in the \textsc{Gemma} family \citep{team2024gemma}, which is observed to reduce the student model's intrinsic entropy and illicit mode-seeking behavior \citep{gu2024minillm,cha2025why}.
    
    \item \textbf{Domain Experiment Scope:} Due to limited compute, we confine the scope of the cross-domain generalization experiment to instruct models under 14B parameters.
    
    \item \textbf{Reward Model Input Formatting:} For the writing quality evaluations, inputs are obtained by concatenating each context--continuation pair, filtering for passages between 150--400 words to align with the length of the reward model's training data.

    \item \textbf{Qualitative Inspection of High-Uncertainty Text:} Beyond the surface-rarity control built into PMI and CPMI, we conducted thorough manual inspection of story continuations with high uncertainty metrics, which generally exhibit a richness in constructive literary uncertainty (figurative language, suspense, formal novelty) rather than confounders such as rare vocabulary, named entities, unusual punctuation, or genre mismatch. This reading is qualitative. We did not have the budget for a formal human evaluation to validate it, and flag that as future work in Appendix~\ref{app:supplemental_discussion}.

    \item \textbf{Reward Model Validation:} The \textsc{Writing Quality Reward Model} \citep{chakrabarty2025aislop} used in Section~\ref{sec:quality} has itself undergone human evaluation against expert creative-writing preferences, which is why we treat its scores as a proxy for writing quality.
\end{itemize}

\section{Statistical Significance of Base to Post-trained Changes}
\label{app:significance}

Section~\ref{sec:results} claims that instruct and thinking variants exhibit larger uncertainty gaps than their base counterparts. This appendix tests that claim for every family in Table~\ref{tab:main_results} with both a base and a post-trained variant: 11 families for base to instruct, 3 for base to thinking.

\paragraph{Method.} Cumulative context--continuation windows within a story are not independent, so we treat each full story as the unit of analysis and take each variant's per-story median. Base and post-trained variants are evaluated on the identical story set, which makes the pairing exact, and we run a paired Wilcoxon signed-rank test across stories. We report the matched-pairs rank-biserial correlation $r_{rb}$ as the effect size, where $|r_{rb}| = 1$ means every story moved in the same direction, and apply Holm--Bonferroni correction across all 168 comparisons. A window-level robustness pass ($n$ up to ${\sim}23$k) yields the same conclusions with smaller p-values. With this many paired observations p-values are tiny almost everywhere, so the effect sizes are what carry the argument.\footnote{This appendix aggregates per story, whereas Table~\ref{tab:main_results} aggregates per window. The two differ by one to two hundredths in the reported medians. Appendix~\ref{app:pmi_cpmi} uses window-level aggregation and reproduces Table~\ref{tab:main_results} exactly.}

\begin{table}[t]
\centering
\small
\caption{Number of family$\times$dataset comparisons that are both significant (Holm-corrected $p<.05$) and in the hypothesized direction, out of 22 for base to instruct and 6 for base to thinking. Expected direction: NLL/PPL ratios increase (the gap widens), $\Delta$PMI and $\Delta$CPMI become more negative, and the model's own NLL/PPL decrease.}
\label{tab:sig_summary}
\begin{tabular}{llccc}
\toprule
\textbf{Target} & \textbf{Comparison} & \textbf{Metric} & \textbf{Sig.\ in expected dir.} & \textbf{Median $|r_{rb}|$} \\
\midrule
Human--model gap & Base$\rightarrow$Instruct & NLL & 18/22 & 1.00 \\
Human--model gap & Base$\rightarrow$Instruct & PPL & 20/22 & 1.00 \\
Human--model gap & Base$\rightarrow$Instruct & PMI & 14/22 & 0.75 \\
Human--model gap & Base$\rightarrow$Instruct & CPMI & 12/22 & 0.80 \\
Human--model gap & Base$\rightarrow$Thinking & NLL & 6/6 & 1.00 \\
Human--model gap & Base$\rightarrow$Thinking & PPL & 6/6 & 1.00 \\
Human--model gap & Base$\rightarrow$Thinking & PMI & 3/6 & 0.85 \\
Human--model gap & Base$\rightarrow$Thinking & CPMI & 2/6 & 0.78 \\
Model's own surprisal & Base$\rightarrow$Instruct & NLL & 18/22 & 1.00 \\
Model's own surprisal & Base$\rightarrow$Instruct & PPL & 18/22 & 1.00 \\
Model's own surprisal & Base$\rightarrow$Thinking & NLL & 6/6 & 1.00 \\
Model's own surprisal & Base$\rightarrow$Thinking & PPL & 6/6 & 1.00 \\
\bottomrule
\end{tabular}
\end{table}

\begin{table*}[t]
\centering
\small
\setlength{\tabcolsep}{4pt}
\caption{Human--model gap, base to instruct. Cells show \emph{base median} $\rightarrow$ \emph{instruct median}; stars denote a significant two-sided change on the Holm-corrected p-value ($^{***}p<.001$, $^{**}p<.01$, $^{*}p<.05$; no star = not significant), and the direction is read from the two medians. NLL/PPL widen for every family except \textsc{Llama-3.1-8B}, which is flat, and \textsc{Mistral-Small-24B}, which narrows slightly.}
\label{tab:sig_gap_instruct}
\begin{tabular}{lcccc}
\toprule
\textbf{Family} & \textbf{NLL} & \textbf{PPL} & \textbf{PMI} & \textbf{CPMI} \\
\midrule
\multicolumn{5}{l}{\textit{New Yorker}} \\
\midrule
OLMo-2-7B         & \shift{2.18}{3.02}$^{***}$ & \shift{3.67}{6.26}$^{***}$ & \shift{-1.71}{-1.87}$^{***}$ & \shift{0.15}{0.03}$^{***}$ \\
OLMo-2-13B        & \shift{2.19}{2.50}$^{***}$ & \shift{3.66}{4.38}$^{***}$ & \shift{-1.64}{-1.80}$^{***}$ & \shift{0.14}{0.13} \\
OLMo-2-32B        & \shift{2.15}{2.32}$^{***}$ & \shift{3.38}{3.71}$^{***}$ & \shift{-1.56}{-1.67}$^{***}$ & \shift{0.11}{0.09} \\
OLMo-3-7B         & \shift{2.20}{3.33}$^{***}$ & \shift{4.10}{8.55}$^{***}$ & \shift{-1.80}{-2.22}$^{***}$ & \shift{0.17}{0.13}$^{***}$ \\
OLMo-3-32B        & \shift{2.16}{3.00}$^{***}$ & \shift{3.52}{5.88}$^{***}$ & \shift{-1.69}{-1.67}        & \shift{0.11}{-0.37}$^{***}$ \\
Llama-3.1-8B      & \shift{2.17}{2.13}         & \shift{3.65}{3.81}$^{*}$   & \shift{-2.43}{-2.58}$^{***}$ & \shift{-0.40}{-0.25}$^{***}$ \\
Qwen3-4B          & \shift{2.22}{3.27}$^{***}$ & \shift{4.16}{7.58}$^{***}$ & \shift{-1.74}{-1.57}$^{***}$ & \shift{0.20}{0.30}$^{***}$ \\
Mistral-Small-24B & \shift{2.13}{2.08}$^{***}$ & \shift{3.32}{3.32}         & \shift{-2.03}{-2.02}         & \shift{-0.48}{-0.56}$^{***}$ \\
Gemma-3-4B        & \shift{2.21}{3.92}$^{***}$ & \shift{3.99}{9.11}$^{***}$ & \shift{-2.01}{-1.97}$^{***}$ & \shift{-0.57}{-1.02}$^{***}$ \\
Gemma-3-12B       & \shift{2.16}{3.27}$^{***}$ & \shift{3.58}{6.51}$^{***}$ & \shift{-1.91}{-1.81}         & \shift{-0.62}{-0.97}$^{***}$ \\
Gemma-3-27B       & \shift{2.34}{3.61}$^{***}$ & \shift{3.81}{6.81}$^{***}$ & \shift{-1.75}{-1.90}$^{***}$ & \shift{-0.79}{-0.84}$^{**}$ \\
\midrule
\multicolumn{5}{l}{\textit{Tell-Me-A-Story}} \\
\midrule
OLMo-2-7B         & \shift{2.16}{3.24}$^{***}$ & \shift{3.16}{5.58}$^{***}$ & \shift{-1.56}{-1.87}$^{***}$ & \shift{0.06}{0.03}$^{***}$ \\
OLMo-2-13B        & \shift{2.16}{2.60}$^{***}$ & \shift{3.09}{3.88}$^{***}$ & \shift{-1.49}{-1.71}$^{***}$ & \shift{0.05}{0.07}$^{**}$ \\
OLMo-2-32B        & \shift{2.14}{2.42}$^{***}$ & \shift{2.94}{3.49}$^{***}$ & \shift{-1.46}{-1.56}$^{***}$ & \shift{0.03}{0.05}$^{***}$ \\
OLMo-3-7B         & \shift{2.15}{3.49}$^{***}$ & \shift{3.36}{7.28}$^{***}$ & \shift{-1.66}{-2.12}$^{***}$ & \shift{0.06}{0.10}$^{**}$ \\
OLMo-3-32B        & \shift{2.12}{3.04}$^{***}$ & \shift{3.03}{4.89}$^{***}$ & \shift{-1.51}{-1.60}$^{***}$ & \shift{0.00}{-0.44}$^{***}$ \\
Llama-3.1-8B      & \shift{2.09}{2.09}         & \shift{3.06}{3.23}$^{***}$ & \shift{-2.34}{-2.47}$^{***}$ & \shift{-0.49}{-0.30}$^{***}$ \\
Qwen3-4B          & \shift{2.18}{3.37}$^{***}$ & \shift{3.52}{6.46}$^{***}$ & \shift{-1.55}{-1.50}$^{***}$ & \shift{0.04}{0.24}$^{***}$ \\
Mistral-Small-24B & \shift{2.07}{2.04}$^{***}$ & \shift{2.81}{2.80}         & \shift{-1.92}{-1.95}$^{**}$  & \shift{-0.54}{-0.62}$^{***}$ \\
Gemma-3-4B        & \shift{2.13}{3.83}$^{***}$ & \shift{3.33}{7.17}$^{***}$ & \shift{-1.96}{-1.92}$^{**}$  & \shift{-0.68}{-1.12}$^{***}$ \\
Gemma-3-12B       & \shift{2.07}{3.23}$^{***}$ & \shift{3.02}{5.14}$^{***}$ & \shift{-1.84}{-1.86}         & \shift{-0.71}{-1.02}$^{***}$ \\
Gemma-3-27B       & \shift{2.26}{3.57}$^{***}$ & \shift{3.18}{5.49}$^{***}$ & \shift{-1.71}{-1.92}$^{***}$ & \shift{-0.82}{-0.83} \\
\bottomrule
\end{tabular}
\end{table*}

\begin{table}[t]
\centering
\small
\caption{Human--model gap, base to thinking, for the three families with an exact base--instruct--thinking triplet. Conventions as in Table~\ref{tab:sig_gap_instruct}.}
\label{tab:sig_gap_thinking}
\begin{tabular}{lcccc}
\toprule
\textbf{Family} & \textbf{NLL} & \textbf{PPL} & \textbf{PMI} & \textbf{CPMI} \\
\midrule
\multicolumn{5}{l}{\textit{New Yorker}} \\
OLMo-3-7B & 2.20$\rightarrow$2.76$^{***}$ & 4.10$\rightarrow$6.19$^{***}$ & $-$1.80$\rightarrow$$-$2.18$^{***}$ & 0.17$\rightarrow$0.15 \\
OLMo-3-32B & 2.16$\rightarrow$2.78$^{***}$ & 3.52$\rightarrow$5.25$^{***}$ & $-$1.69$\rightarrow$$-$1.66 & 0.11$\rightarrow$$-$0.36$^{***}$ \\
Qwen3-4B & 2.22$\rightarrow$3.35$^{***}$ & 4.16$\rightarrow$8.37$^{***}$ & $-$1.74$\rightarrow$$-$1.56$^{***}$ & 0.20$\rightarrow$0.28$^{***}$ \\
\midrule
\multicolumn{5}{l}{\textit{Tell-Me-A-Story}} \\
OLMo-3-7B & 2.15$\rightarrow$2.76$^{***}$ & 3.36$\rightarrow$5.15$^{***}$ & $-$1.66$\rightarrow$$-$2.03$^{***}$ & 0.06$\rightarrow$0.05 \\
OLMo-3-32B & 2.12$\rightarrow$2.81$^{***}$ & 3.03$\rightarrow$4.43$^{***}$ & $-$1.51$\rightarrow$$-$1.57$^{***}$ & 0.00$\rightarrow$$-$0.39$^{***}$ \\
Qwen3-4B & 2.18$\rightarrow$3.38$^{***}$ & 3.52$\rightarrow$6.79$^{***}$ & $-$1.55$\rightarrow$$-$1.44$^{***}$ & 0.04$\rightarrow$0.20$^{***}$ \\
\bottomrule
\end{tabular}
\end{table}

\begin{table}[t]
\centering
\small
\caption{The model continuation's own surprisal, base to instruct, which is the most literal reading of a ``reduction in uncertainty''. It drops significantly for 9 of 11 families, with every story moving the same way ($|r_{rb}| \approx 1$). The two exceptions, \textsc{Llama-3.1-8B} and \textsc{Mistral-Small-24B}, show a small but significant \emph{increase}: post-training did not reduce their uncertainty.}
\label{tab:sig_model_surprisal}
\begin{tabular}{lcccc}
\toprule
& \multicolumn{2}{c}{\textbf{New Yorker}} & \multicolumn{2}{c}{\textbf{Tell-Me-A-Story}} \\
\cmidrule(lr){2-3} \cmidrule(lr){4-5}
\textbf{Family} & \textbf{NLL} & \textbf{PPL} & \textbf{NLL} & \textbf{PPL} \\
\midrule
OLMo-2-7B & 1.11$\rightarrow$0.91$^{***}$ & 3.03$\rightarrow$2.48$^{***}$ & 1.01$\rightarrow$0.78$^{***}$ & 2.76$\rightarrow$2.17$^{***}$ \\
OLMo-2-13B & 1.07$\rightarrow$0.98$^{***}$ & 2.91$\rightarrow$2.67$^{***}$ & 0.98$\rightarrow$0.86$^{***}$ & 2.66$\rightarrow$2.36$^{***}$ \\
OLMo-2-32B & 1.04$\rightarrow$0.97$^{***}$ & 2.82$\rightarrow$2.64$^{***}$ & 0.95$\rightarrow$0.88$^{***}$ & 2.60$\rightarrow$2.41$^{***}$ \\
OLMo-3-7B & 1.16$\rightarrow$0.91$^{***}$ & 3.21$\rightarrow$2.48$^{***}$ & 1.07$\rightarrow$0.80$^{***}$ & 2.90$\rightarrow$2.23$^{***}$ \\
OLMo-3-32B & 1.09$\rightarrow$0.88$^{***}$ & 2.98$\rightarrow$2.41$^{***}$ & 1.00$\rightarrow$0.79$^{***}$ & 2.73$\rightarrow$2.20$^{***}$ \\
Llama-3.1-8B & 1.10$\rightarrow$1.21$^{***}$ & 3.01$\rightarrow$3.35$^{***}$ & 1.02$\rightarrow$1.11$^{***}$ & 2.77$\rightarrow$3.03$^{***}$ \\
Qwen3-4B & 1.19$\rightarrow$0.89$^{***}$ & 3.27$\rightarrow$2.43$^{***}$ & 1.09$\rightarrow$0.80$^{***}$ & 2.96$\rightarrow$2.23$^{***}$ \\
Mistral-Small-24B & 1.06$\rightarrow$1.10$^{***}$ & 2.87$\rightarrow$3.00$^{***}$ & 0.98$\rightarrow$1.01$^{***}$ & 2.66$\rightarrow$2.74$^{***}$ \\
Gemma-3-4B & 1.14$\rightarrow$0.76$^{***}$ & 3.14$\rightarrow$2.14$^{***}$ & 1.07$\rightarrow$0.70$^{***}$ & 2.93$\rightarrow$2.01$^{***}$ \\
Gemma-3-12B & 1.11$\rightarrow$0.81$^{***}$ & 3.04$\rightarrow$2.26$^{***}$ & 1.03$\rightarrow$0.74$^{***}$ & 2.81$\rightarrow$2.09$^{***}$ \\
Gemma-3-27B & 0.99$\rightarrow$0.74$^{***}$ & 2.69$\rightarrow$2.10$^{***}$ & 0.92$\rightarrow$0.66$^{***}$ & 2.50$\rightarrow$1.93$^{***}$ \\
\bottomrule
\end{tabular}
\end{table}

\paragraph{Findings.} Instruction tuning significantly widens the human--model NLL/PPL gap in 18 to 20 of 22 family$\times$dataset cells, and the model continuation's own NLL/PPL drops significantly in 18 of 22 (Tables~\ref{tab:sig_summary},~\ref{tab:sig_gap_instruct},~\ref{tab:sig_model_surprisal}). The two exceptions are \textsc{Llama-3.1-8B} and \textsc{Mistral-Small-24B}, which Section~\ref{sec:results} already identifies as the families exhibiting almost no alignment tax; the test therefore strengthens the existing narrative rather than qualifying it. Thinking variants widen the gap in all 6 of 6 cells (Table~\ref{tab:sig_gap_thinking}). The context-dependence metrics also change significantly, but with family-dependent sign: \textsc{Qwen3-4B}'s PMI gap narrows while \textsc{OLMo-3-32B}'s CPMI gap widens, for instance. Appendix~\ref{app:pmi_cpmi} accounts for this heterogeneity.

\section{Within-family Scaling}
\label{app:scale}

Three families in Table~\ref{tab:main_results} have at least two sizes: \textsc{OLMo-2} \{7B, 13B, 32B\}, \textsc{OLMo-3} \{7B, 32B\}, and \textsc{Gemma-3} \{4B, 12B, 27B\}. This appendix tests the two halves of the scale claim in Section~\ref{sec:results}, namely that scale barely moves the base-model gap but does shrink the alignment tax.

\paragraph{Method.} As in Appendix~\ref{app:significance}, the story is the unit of analysis and we take each variant's per-story median. Within a family and variant, all sizes are evaluated on the identical story set, so pairing across sizes is exact. The primary test is a paired Wilcoxon signed-rank test between the extreme sizes; adjacent-size tests give the same picture and are omitted here. Holm--Bonferroni is applied across the primary tests within each suite.\footnote{Almost every base-model test registers as significant despite a near-zero magnitude. This reflects the power of paired tests at $n = 70$--$120$ stories rather than a substantive scale effect, which is why the closing paragraph reads magnitudes rather than stars.}

\begin{table*}[t]
\centering
\small
\caption{Per-variant scale effect. Cells show the median at the smallest size $\rightarrow$ the median at the largest size, with stars on the Holm-corrected p-value ($^{***}p<.001$, $^{**}p<.01$, $^{*}p<.05$) and the matched-pairs rank-biserial effect size. \textsc{Gemma-3} Base is the one case where the NLL ratio grows slightly with scale.}
\label{tab:scale_per_variant}
\resizebox{\textwidth}{!}{%
\begin{tabular}{llcccc}
\toprule
\textbf{Family} & \textbf{Variant} & \textbf{NLL ratio} & \textbf{PPL ratio} & \textbf{PMI gap} & \textbf{CPMI gap} \\
\midrule
\multicolumn{6}{l}{\textit{New Yorker}} \\
OLMo-2 & Base & 2.18 $\rightarrow$ 2.15$^{*}$ {\scriptsize $r_{rb}\!=\!-.40$} & 3.67 $\rightarrow$ 3.38$^{***}$ {\scriptsize $r_{rb}\!=\!-1.0$} & $-$1.71 $\rightarrow$ $-$1.56$^{***}$ {\scriptsize $r_{rb}\!=\!+.84$} & +0.15 $\rightarrow$ +0.11$^{*}$ {\scriptsize $r_{rb}\!=\!-.43$} \\
OLMo-2 & Instruct & 3.02 $\rightarrow$ 2.32$^{***}$ {\scriptsize $r_{rb}\!=\!-1.0$} & 6.26 $\rightarrow$ 3.71$^{***}$ {\scriptsize $r_{rb}\!=\!-1.0$} & $-$1.87 $\rightarrow$ $-$1.67$^{***}$ {\scriptsize $r_{rb}\!=\!+.95$} & +0.03 $\rightarrow$ +0.09$^{***}$ {\scriptsize $r_{rb}\!=\!+.64$} \\
\addlinespace
OLMo-3 & Base & 2.20 $\rightarrow$ 2.16$^{**}$ {\scriptsize $r_{rb}\!=\!-.47$} & 4.10 $\rightarrow$ 3.52$^{***}$ {\scriptsize $r_{rb}\!=\!-1.0$} & $-$1.80 $\rightarrow$ $-$1.69$^{***}$ {\scriptsize $r_{rb}\!=\!+.85$} & +0.17 $\rightarrow$ +0.11$^{***}$ {\scriptsize $r_{rb}\!=\!-.69$} \\
OLMo-3 & Instruct & 3.33 $\rightarrow$ 3.00$^{***}$ {\scriptsize $r_{rb}\!=\!-1.0$} & 8.55 $\rightarrow$ 5.88$^{***}$ {\scriptsize $r_{rb}\!=\!-1.0$} & $-$2.22 $\rightarrow$ $-$1.67$^{***}$ {\scriptsize $r_{rb}\!=\!+1.0$} & +0.13 $\rightarrow$ $-$0.37$^{***}$ {\scriptsize $r_{rb}\!=\!-1.0$} \\
OLMo-3 & Thinking & 2.76 $\rightarrow$ 2.78 {\scriptsize (ns)} & 6.19 $\rightarrow$ 5.25$^{***}$ {\scriptsize $r_{rb}\!=\!-.97$} & $-$2.18 $\rightarrow$ $-$1.66$^{***}$ {\scriptsize $r_{rb}\!=\!+1.0$} & +0.15 $\rightarrow$ $-$0.36$^{***}$ {\scriptsize $r_{rb}\!=\!-1.0$} \\
\addlinespace
Gemma-3 & Base & 2.21 $\rightarrow$ 2.34$^{***}$ {\scriptsize $r_{rb}\!=\!+.97$} & 3.99 $\rightarrow$ 3.81$^{***}$ {\scriptsize $r_{rb}\!=\!-.84$} & $-$2.01 $\rightarrow$ $-$1.75$^{***}$ {\scriptsize $r_{rb}\!=\!+1.0$} & $-$0.57 $\rightarrow$ $-$0.79$^{***}$ {\scriptsize $r_{rb}\!=\!-1.0$} \\
Gemma-3 & Instruct & 3.92 $\rightarrow$ 3.61$^{***}$ {\scriptsize $r_{rb}\!=\!-.93$} & 9.11 $\rightarrow$ 6.81$^{***}$ {\scriptsize $r_{rb}\!=\!-1.0$} & $-$1.97 $\rightarrow$ $-$1.90 {\scriptsize (ns)} & $-$1.02 $\rightarrow$ $-$0.84$^{***}$ {\scriptsize $r_{rb}\!=\!+.93$} \\
\midrule
\multicolumn{6}{l}{\textit{Tell-Me-A-Story}} \\
OLMo-2 & Base & 2.16 $\rightarrow$ 2.14 {\scriptsize (ns)} & 3.16 $\rightarrow$ 2.94$^{***}$ {\scriptsize $r_{rb}\!=\!-.93$} & $-$1.56 $\rightarrow$ $-$1.46$^{***}$ {\scriptsize $r_{rb}\!=\!+.73$} & +0.06 $\rightarrow$ +0.03$^{***}$ {\scriptsize $r_{rb}\!=\!-.49$} \\
OLMo-2 & Instruct & 3.24 $\rightarrow$ 2.42$^{***}$ {\scriptsize $r_{rb}\!=\!-1.0$} & 5.58 $\rightarrow$ 3.49$^{***}$ {\scriptsize $r_{rb}\!=\!-1.0$} & $-$1.87 $\rightarrow$ $-$1.56$^{***}$ {\scriptsize $r_{rb}\!=\!+.99$} & +0.03 $\rightarrow$ +0.05$^{*}$ {\scriptsize $r_{rb}\!=\!+.29$} \\
\addlinespace
OLMo-3 & Base & 2.15 $\rightarrow$ 2.12$^{***}$ {\scriptsize $r_{rb}\!=\!-.49$} & 3.36 $\rightarrow$ 3.03$^{***}$ {\scriptsize $r_{rb}\!=\!-.97$} & $-$1.66 $\rightarrow$ $-$1.51$^{***}$ {\scriptsize $r_{rb}\!=\!+.94$} & +0.06 $\rightarrow$ +0.00$^{***}$ {\scriptsize $r_{rb}\!=\!-.68$} \\
OLMo-3 & Instruct & 3.49 $\rightarrow$ 3.04$^{***}$ {\scriptsize $r_{rb}\!=\!-.98$} & 7.28 $\rightarrow$ 4.89$^{***}$ {\scriptsize $r_{rb}\!=\!-1.0$} & $-$2.12 $\rightarrow$ $-$1.60$^{***}$ {\scriptsize $r_{rb}\!=\!+1.0$} & +0.10 $\rightarrow$ $-$0.44$^{***}$ {\scriptsize $r_{rb}\!=\!-1.0$} \\
OLMo-3 & Thinking & 2.76 $\rightarrow$ 2.81$^{***}$ {\scriptsize $r_{rb}\!=\!+.48$} & 5.15 $\rightarrow$ 4.44$^{***}$ {\scriptsize $r_{rb}\!=\!-.95$} & $-$2.03 $\rightarrow$ $-$1.57$^{***}$ {\scriptsize $r_{rb}\!=\!+1.0$} & +0.05 $\rightarrow$ $-$0.39$^{***}$ {\scriptsize $r_{rb}\!=\!-1.0$} \\
\addlinespace
Gemma-3 & Base & 2.13 $\rightarrow$ 2.26$^{***}$ {\scriptsize $r_{rb}\!=\!+.96$} & 3.33 $\rightarrow$ 3.18$^{***}$ {\scriptsize $r_{rb}\!=\!-.85$} & $-$1.96 $\rightarrow$ $-$1.71$^{***}$ {\scriptsize $r_{rb}\!=\!+1.0$} & $-$0.68 $\rightarrow$ $-$0.82$^{***}$ {\scriptsize $r_{rb}\!=\!-.97$} \\
Gemma-3 & Instruct & 3.83 $\rightarrow$ 3.57$^{***}$ {\scriptsize $r_{rb}\!=\!-.87$} & 7.17 $\rightarrow$ 5.49$^{***}$ {\scriptsize $r_{rb}\!=\!-1.0$} & $-$1.92 $\rightarrow$ $-$1.92 {\scriptsize (ns)} & $-$1.12 $\rightarrow$ $-$0.83$^{***}$ {\scriptsize $r_{rb}\!=\!+1.0$} \\
\bottomrule
\end{tabular}
}
\end{table*}

\begin{table*}[t]
\centering
\small
\caption{Alignment tax by scale, where the tax is the instruct gap minus the base gap. Cells show the per-story tax at the smallest size $\rightarrow$ the tax at the largest size. All 24 tests are significant at Holm $p<.001$.}
\label{tab:scale_tax}
\resizebox{\textwidth}{!}{%
\begin{tabular}{lcccc}
\toprule
\textbf{Family} & \textbf{NLL tax} & \textbf{PPL tax} & \textbf{PMI tax} & \textbf{CPMI tax} \\
\midrule
\multicolumn{5}{l}{\textit{New Yorker}} \\
OLMo-2 (7B$\rightarrow$32B) & +0.83 $\rightarrow$ +0.16$^{***}$ {\scriptsize $r_{rb}\!=\!-1.0$} & +2.44 $\rightarrow$ +0.34$^{***}$ {\scriptsize $r_{rb}\!=\!-1.0$} & $-$0.19 $\rightarrow$ $-$0.10$^{***}$ {\scriptsize $r_{rb}\!=\!+.50$} & $-$0.11 $\rightarrow$ $-$0.01$^{***}$ {\scriptsize $r_{rb}\!=\!+.78$} \\
OLMo-3 (7B$\rightarrow$32B) & +1.12 $\rightarrow$ +0.86$^{***}$ {\scriptsize $r_{rb}\!=\!-.98$} & +4.22 $\rightarrow$ +2.22$^{***}$ {\scriptsize $r_{rb}\!=\!-1.0$} & $-$0.38 $\rightarrow$ $-$0.03$^{***}$ {\scriptsize $r_{rb}\!=\!+.98$} & $-$0.04 $\rightarrow$ $-$0.47$^{***}$ {\scriptsize $r_{rb}\!=\!-1.0$} \\
Gemma-3 (4B$\rightarrow$27B) & +1.72 $\rightarrow$ +1.26$^{***}$ {\scriptsize $r_{rb}\!=\!-.98$} & +5.15 $\rightarrow$ +2.92$^{***}$ {\scriptsize $r_{rb}\!=\!-1.0$} & +0.10 $\rightarrow$ $-$0.15$^{***}$ {\scriptsize $r_{rb}\!=\!-.92$} & $-$0.46 $\rightarrow$ $-$0.06$^{***}$ {\scriptsize $r_{rb}\!=\!+1.0$} \\
\midrule
\multicolumn{5}{l}{\textit{Tell-Me-A-Story}} \\
OLMo-2 (7B$\rightarrow$32B) & +1.08 $\rightarrow$ +0.28$^{***}$ {\scriptsize $r_{rb}\!=\!-1.0$} & +2.39 $\rightarrow$ +0.49$^{***}$ {\scriptsize $r_{rb}\!=\!-1.0$} & $-$0.29 $\rightarrow$ $-$0.10$^{***}$ {\scriptsize $r_{rb}\!=\!+.84$} & $-$0.03 $\rightarrow$ +0.02$^{***}$ {\scriptsize $r_{rb}\!=\!+.53$} \\
OLMo-3 (7B$\rightarrow$32B) & +1.36 $\rightarrow$ +0.91$^{***}$ {\scriptsize $r_{rb}\!=\!-.97$} & +3.88 $\rightarrow$ +1.87$^{***}$ {\scriptsize $r_{rb}\!=\!-1.0$} & $-$0.44 $\rightarrow$ $-$0.11$^{***}$ {\scriptsize $r_{rb}\!=\!+.96$} & +0.03 $\rightarrow$ $-$0.47$^{***}$ {\scriptsize $r_{rb}\!=\!-1.0$} \\
Gemma-3 (4B$\rightarrow$27B) & +1.66 $\rightarrow$ +1.31$^{***}$ {\scriptsize $r_{rb}\!=\!-.97$} & +3.70 $\rightarrow$ +2.32$^{***}$ {\scriptsize $r_{rb}\!=\!-1.0$} & +0.05 $\rightarrow$ $-$0.21$^{***}$ {\scriptsize $r_{rb}\!=\!-.95$} & $-$0.44 $\rightarrow$ $-$0.02$^{***}$ {\scriptsize $r_{rb}\!=\!+1.0$} \\
\bottomrule
\end{tabular}
}
\end{table*}

\paragraph{Findings.} The base-model gap is practically flat with scale. Base NLL ratios change by at most $0.13$ between extreme sizes, against instruct NLL ratios that change by $0.26$ to $0.82$ (Table~\ref{tab:scale_per_variant}). The alignment tax, by contrast, shrinks monotonically with scale for NLL and PPL in every family on both datasets, all 12 of 12 tests significant at Holm $p<.001$ with $|r_{rb}| \approx 1$ (Table~\ref{tab:scale_tax}). For \textsc{OLMo-2} on \textit{New Yorker} the NLL tax falls from $+0.83$ at 7B to $+0.16$ at 32B, a fivefold reduction, and the 38\% to 8\% figure cited in Section~\ref{sec:results} is now joined by the same pattern in \textsc{OLMo-3} and \textsc{Gemma-3}.

Three qualifications are worth stating plainly. \textsc{Gemma-3} Base is the one variant whose NLL ratio \emph{grows} with scale, by $+0.13$ on both datasets, though this is small in absolute terms. \textsc{Gemma-3} Instruct's PMI is the single non-significant cell in Table~\ref{tab:scale_per_variant}. And the \textsc{OLMo-3} CPMI tax \emph{grows} rather than shrinks with scale, which the decomposition in Appendix~\ref{app:pmi_cpmi} predicts: post-training at 32B shifts that family's gap onto decision points. \textsc{Gemma-3} Instruct NLL is also non-monotone across 4B, 12B, and 27B (3.90, 3.29, 3.60), so the defensible claim is about the end-to-end direction rather than a clean slope.

\section{Prompt-time Interventions}
\label{app:prompting}

Since post-training widens the gap, a natural question is whether prompt-time methods can narrow it. We consider two recent proposals, Verbalized Sampling \citep[VS;][]{zhang2025verbalized} and NeoCoder \citep{lu2025rethinking}, and report two supplemental probes that fit our measurement frame.

\paragraph{Why the suggested methods do not drop in.} Our pipeline is a pure continuation: model and human continuations are scored under identical conditioning on the raw prefix $T_1$, with no chat template, system prompt, or user instruction. VS deliberately samples from a verbalized JSON distribution constructed to differ from the model's softmax, so a VS continuation is not drawn from $P_\theta(\cdot \mid T_1)$, the object Table~\ref{tab:main_results} measures. VS also requires trained chat and JSON behavior, so it cannot run on base models at all, which rules out the base-to-instruct contrast this paper turns on. NeoCoder is a domain mismatch: it relies on the verifiable ``obvious-solution'' structure of code, and adapting denial or constraint prompting to fiction would be a separate study.

\paragraph{Probes.} Both run on a deterministic 20\% story sample drawn with seed 42 after the same 4096-token filter as the main experiment, giving 1,709 \textit{New Yorker} and 9,103 \textit{Tell-Me-A-Story} pairs, with one instruct model per family at 8B or below. The \emph{temperature sweep} varies $\texttt{temperature} \in \{0.6, 1.0, 1.5\}$, a drop-in change to \texttt{SamplingParams} that preserves the bare-$T_1$ measurement frame and is therefore directly comparable to the Instruct rows of Table~\ref{tab:main_results}; the $T = 1.0$ row is recomputed from our main per-pair results restricted to the same sample. The \emph{scoped VS probe} generates $K = 5$ candidates with verbalized probabilities, samples one, truncates it to $|T_2|$ tokens, and teacher-force-scores it under bare $T_1$, so the measurement frame matches Table~\ref{tab:main_results} while the generative frame differs. The VS probe completed for \textsc{OLMo-2-7B-Instruct} and \textsc{Llama-3.1-8B-Instruct} only; the \textsc{Qwen3-4B} and \textsc{Gemma-3-4B} cells were not run.

\begin{table}[t]
\centering
\small
\caption{Median paired gap under bare-$T_1$ scoring, reported per dataset as New Yorker / Tell-Me-A-Story. Note that the sign convention differs from Table~\ref{tab:main_results}: these are gaps in nats, not ratios. A positive NLL gap means the model is more confident than the human; a negative PMI gap means model continuations depend on context more than human ones. PPL ratio equals $\exp$(NLL gap) and tracks NLL directly.}
\label{tab:prompting}
\begin{tabular}{lcccc}
\toprule
\textbf{Setting} & \textbf{OLMo-2-7B} & \textbf{Llama-3.1-8B} & \textbf{Qwen3-4B} & \textbf{Gemma-3-4B} \\
\midrule
\multicolumn{5}{l}{\textit{Gap in NLL (nats)}} \\
$T=0.6$ & 1.87 / 1.81 & 1.38 / 1.22 & 2.00 / 1.83 & 2.23 / 1.96 \\
$T=1.0$ & 1.81 / 1.72 & 1.33 / 1.17 & 2.02 / 1.86 & 2.23 / 1.96 \\
$T=1.5$ & 1.67 / 1.63 & $-$0.33 / $-$0.29 & 1.81 / 1.67 & 2.08 / 1.82 \\
VS & $-$0.10 / $-$0.25 & $-$0.40 / $-$0.27 & - & - \\
\midrule
\multicolumn{5}{l}{\textit{Gap in PMI (nats)}} \\
$T=0.6$ & $-$1.42 / $-$1.53 & $-$1.63 / $-$1.68 & $-$0.82 / $-$0.87 & $-$1.33 / $-$1.41 \\
$T=1.0$ & $-$1.87 / $-$1.87 & $-$2.61 / $-$2.50 & $-$1.56 / $-$1.47 & $-$1.94 / $-$1.94 \\
$T=1.5$ & $-$5.31 / $-$4.78 & $-$5.79 / $-$5.71 & $-$4.72 / $-$4.11 & $-$5.10 / $-$4.73 \\
VS & $+$0.45 / $+$0.37 & $+$0.49 / $+$0.45 & - & - \\
\bottomrule
\end{tabular}
\end{table}

\paragraph{Temperature has little effect.} For 3 of the 4 models the NLL gap is essentially insensitive to $T$, remaining at or above 1.6 nats across the entire range (Table~\ref{tab:prompting}). The one apparent exception, \textsc{Llama-3.1-8B} at $T = 1.5$, closes the gap only by degenerating: manual inspection of its outputs shows clear incoherence, and the model's own PPL exceeds the human's. The PMI gap moves consistently in the opposite direction across all four models, growing roughly three times more negative as $T$ rises from 1.0 to 1.5, so model continuations become \emph{more} PMI-rich relative to humans, the opposite of human-like.

\paragraph{VS shifts the measured gap, but measures a different object.} On the parsed subset the NLL gap collapses to slightly negative and the PMI gap flips sign. Three caveats limit the reading. First, the chosen continuation is drawn from the verbalized distribution rather than $P_\theta(\cdot \mid T_1)$, so ``closing the gap'' here is not the quantity Table~\ref{tab:main_results} reports. Second, parse-failure rates for the $K=5$ JSON instruction are high, at 68\% and 69\% for \textsc{OLMo-2} and 36\% and 35\% for \textsc{Llama-3.1} on \textit{New Yorker} and \textit{Tell-Me-A-Story} respectively, so the VS rows cover a subset of the temperature sweep. Third, conditioning is structurally different, since the candidate was generated under a chat template with a system prompt and the VS instruction rather than bare $T_1$. The PMI sign flip in particular suggests that VS selects lower-PMI candidates from the verbalized list, not that $P_\theta(\cdot \mid T_1)$ has itself become more human-like.

Prompt-time interventions can therefore shift the measured gap, VS substantially and temperature only through degeneration, but neither undoes the central finding that post-training widens the gap relative to the base model under matched conditioning on bare $T_1$. The base-model rows of Table~\ref{tab:main_results} remain the strongest existence proof that the gap is avoidable through training-time choices rather than decoding or prompting.

\section{Does Uncertainty Correlate with Writing Quality?}
\label{app:quality_detailed}

This section presents comprehensive results for the correlation and regression analyses
between uncertainty metrics and writing quality scores described in Section~\ref{sec:quality}.

\subsection{Interpreting the Effect Sizes}
\label{app:quality_effect_sizes}

The correlations in Section~\ref{sec:quality} are modest in absolute terms: $\bar{\rho} \approx 0.07$--$0.11$ for NLL and PPL, and $-0.08$ to $-0.11$ for PMI. This subsection sets out what they are and are not meant to establish.

We do not claim that uncertainty \emph{drives} writing quality. That is a causal claim, and it would require interventions on uncertainty such as decoding-time perturbations, which we discuss as future work. Section~\ref{sec:quality} asks the narrower question of whether uncertainty is correlated with writing quality at all, and it serves mainly as a check on Table~\ref{tab:main_results}: it tests whether the high uncertainty of human text reflects noise or incoherence rather than a proxy measurement of literary quality. Two considerations make these correlations the right kind of evidence for that check.

First, the direction and consistency of the correlations carry the argument, not their magnitude. Across 28 models and 3 datasets, two independent metrics agree on the direction that literary theory predicts (Section~\ref{sec:theory}): higher token-level uncertainty tracks higher quality, and higher context-dependence tracks lower quality. The structure is interpretable beyond sign, since quadratic regression locates sweet-spot peaks at $\bar{z}^* = 1.98$ standard deviations above mean uncertainty for human text, a moderately high level of novelty before incoherence sets in, as the Wundt curve predicts. Noise would not produce this kind of cross-model, cross-metric, and theory-consistent structure.

Second, the effect sizes are what one should expect from a single ingredient of a multi-dimensional construct. Writing quality is shaped by plot, character, voice, thematic coherence, and further dimensions that uncertainty does not measure, and Appendix~\ref{app:supplemental_discussion} notes that uncertainty is a necessary but likely not sufficient condition for literary quality. A modest but consistently signed correlation with one of many quality ingredients is the predicted pattern under that framing; a large $\rho$ would instead raise the suspicion that the quality scorer is conflating quality with surprisal. The dataset-level heterogeneity supports this reading. \textit{New Yorker} correlations are near zero ($\rho \approx 0.02$), which fits a ceiling-effect interpretation in which the highest-quality stories already sit near the sweet-spot peak, while \textit{Tell-Me-A-Story} and \textit{Ghostbuster} have more variance in writing quality and show a considerably larger $\rho$.

\subsection{Per-Dataset Breakdown}

Table~\ref{tab:quality_detailed} presents the full correlation statistics disaggregated by dataset. 
Several patterns emerge from this breakdown. First, the positive correlation between NLL/PPL and 
quality is strongest in \textit{Tell-Me-A-Story} (human: $\rho = 0.108$; model: $\rho = 0.146$) and 
\textit{Ghostbuster} (human: $\rho = 0.122$; model: $\rho = 0.176$), with 100\% of models showing 
significant positive correlations in both datasets. In contrast, \textit{New Yorker} exhibits weaker, 
largely non-significant correlations ($\rho \approx 0.02$ for human text), potentially reflecting 
the more heterogeneous nature of magazine writing or ceiling effects in the quality scores for 
professionally edited prose.

Second, the sweet-spot phenomenon is dataset-dependent. For human-authored text, 79\% of models 
detect an inverted-U relationship in \textit{Tell-Me-A-Story}, compared to 0\% in \textit{New Yorker} 
and \textit{Ghostbuster}. The mean peak location for these sweet-spot cases occurs at $z^* = 1.98$ 
standard deviations above the mean uncertainty, suggesting that optimal quality is associated with 
text that is substantially more surprising than average, but not maximally so.

Third, the PMI metric consistently shows negative correlations across all datasets, with the strongest 
effect in \textit{Tell-Me-A-Story} (human: $\rho = -0.160$, 100\% significantly negative). This confirms 
that high contextual predictability—where the preceding context strongly determines upcoming tokens—is 
associated with lower quality scores, possibly because such text relies on clichéd or formulaic constructions.

\begin{table*}[t]
\centering
\small
\caption{Detailed correlation analysis between uncertainty metrics and writing quality, broken down by dataset. $\bar{\rho}$: mean Spearman correlation across models; $\sigma_\rho$: standard deviation; Sig.+/--: percentage of models with significant positive/negative correlation ($p < 0.05$); Sweet: percentage exhibiting inverted-U relationship; $\bar{z}^*$: mean peak location in standardized units for sweet-spot cases.}
\label{tab:quality_detailed}
\begin{tabular}{llccccccccc}
\toprule
\textbf{Dataset} & \textbf{Source} & \textbf{Metric} & $n$ & $\bar{\rho}$ & $\sigma_\rho$ & \textbf{Sig.+} & \textbf{Sig.--} & \textbf{Sweet} & $\bar{z}^*$ & $\Delta R^2$ \\
\midrule
ghostbuster & Human & NLL & 8 & 0.122 & 0.021 & 100\% & 0\% & 0\% & -- & 0.0008 \\
 & Human & PMI & 8 & -0.077 & 0.025 & 0\% & 88\% & 12\% & -0.68 & 0.0007 \\
 & Human & PPL & 8 & 0.122 & 0.021 & 100\% & 0\% & 0\% & -- & 0.0008 \\
 & Model & NLL & 8 & 0.176 & 0.033 & 100\% & 0\% & 0\% & -- & 0.0016 \\
 & Model & PMI & 8 & -0.031 & 0.030 & 0\% & 25\% & 0\% & -- & 0.0009 \\
 & Model & PPL & 8 & 0.176 & 0.033 & 100\% & 0\% & 0\% & -- & 0.0046 \\
\midrule
new\_yorker & Human & NLL & 28 & 0.021 & 0.012 & 0\% & 0\% & 0\% & -- & 0.0003 \\
 & Human & PMI & 28 & -0.060 & 0.015 & 0\% & 46\% & 0\% & -- & 0.0005 \\
 & Human & PPL & 28 & 0.021 & 0.012 & 0\% & 0\% & 0\% & -- & 0.0007 \\
 & Model & NLL & 28 & 0.053 & 0.040 & 46\% & 0\% & 29\% & 0.80 & 0.0027 \\
 & Model & PMI & 28 & -0.057 & 0.030 & 0\% & 50\% & 4\% & 1.75 & 0.0016 \\
 & Model & PPL & 28 & 0.053 & 0.040 & 46\% & 0\% & 21\% & 1.31 & 0.0025 \\
\midrule
tell\_me\_a\_story & Human & NLL & 28 & 0.108 & 0.025 & 100\% & 0\% & 79\% & 1.98 & 0.0034 \\
 & Human & PMI & 28 & -0.160 & 0.021 & 0\% & 100\% & 0\% & -- & 0.0018 \\
 & Human & PPL & 28 & 0.108 & 0.025 & 100\% & 0\% & 0\% & -- & 0.0011 \\
 & Model & NLL & 28 & 0.146 & 0.031 & 100\% & 0\% & 43\% & 1.92 & 0.0027 \\
 & Model & PMI & 28 & -0.108 & 0.025 & 0\% & 100\% & 0\% & -- & 0.0005 \\
 & Model & PPL & 28 & 0.146 & 0.031 & 100\% & 0\% & 50\% & 2.12 & 0.0071 \\
\bottomrule
\end{tabular}
\end{table*}

\subsection{Per-Model Results}

Table~\ref{tab:quality_per_model} shows Spearman correlations between NLL and writing quality for 
each individual model. The results demonstrate remarkable consistency across model families: nearly 
all models show significant positive correlations for \textit{Tell-Me-A-Story} and \textit{Ghostbuster}, 
while \textit{New Yorker} correlations are uniformly weaker. This consistency suggests that the 
uncertainty–quality relationship is a robust property of the text rather than an artifact of any 
particular model's probability estimates.

Among the models evaluated, instruction-tuned variants (e.g., \textsc{gemma-3-4b-it}, 
\textsc{OLMo-2-1124-7B-Instruct}) tend to show stronger correlations for model-generated text 
than their base counterparts, likely because instruction tuning shifts the probability mass in 
ways that make quality-correlated deviations more detectable. The strongest individual correlation 
observed is $\rho = 0.218$ (\textsc{gemma-3-12b-it} on \textit{Tell-Me-A-Story} model continuations).

\begin{table*}[t]
\centering
\footnotesize
\caption{Per-model Spearman correlations between uncertainty (NLL) and writing quality. Significance: $^{***}p<0.001$, $^{**}p<0.01$, $^{*}p<0.05$.}
\label{tab:quality_per_model}
\begin{tabular}{lcccccc}
\toprule
& \multicolumn{2}{c}{\textbf{New Yorker}} & \multicolumn{2}{c}{\textbf{TellMeAStory}} & \multicolumn{2}{c}{\textbf{Ghostbuster}} \\
\cmidrule(lr){2-3} \cmidrule(lr){4-5} \cmidrule(lr){6-7}
\textbf{Model} & Human & Model & Human & Model & Human & Model \\
\midrule
Qwen3-4B-Base & 0.029 & 0.018 & 0.130$^{***}$ & 0.121$^{***}$ & -- & -- \\
Qwen3-4B-Instruct-2507 & 0.015 & 0.100$^{**}$ & 0.095$^{***}$ & 0.167$^{***}$ & 0.112$^{***}$ & 0.176$^{***}$ \\
Qwen3-4B-Thinking-2507 & 0.017 & 0.064$^{*}$ & 0.090$^{***}$ & 0.109$^{***}$ & -- & -- \\
OLMo-2-0325-32B & 0.007 & 0.017 & 0.134$^{***}$ & 0.153$^{***}$ & -- & -- \\
OLMo-2-0325-32B-Instruct & 0.006 & 0.026 & 0.111$^{***}$ & 0.162$^{***}$ & -- & -- \\
OLMo-2-1124-13B & 0.016 & -0.016 & 0.130$^{***}$ & 0.114$^{***}$ & -- & -- \\
OLMo-2-1124-13B-Instruct & 0.014 & 0.013 & 0.107$^{***}$ & 0.168$^{***}$ & 0.131$^{***}$ & 0.203$^{***}$ \\
OLMo-2-1124-7B & 0.006 & 0.054 & 0.121$^{***}$ & 0.117$^{***}$ & -- & -- \\
OLMo-2-1124-7B-Instruct & 0.011 & 0.070$^{*}$ & 0.076$^{***}$ & 0.180$^{***}$ & 0.091$^{***}$ & 0.213$^{***}$ \\
Olmo-3-1025-7B & 0.032 & 0.074$^{*}$ & 0.113$^{***}$ & 0.140$^{***}$ & -- & -- \\
Olmo-3-1125-32B & 0.014 & 0.059 & 0.126$^{***}$ & 0.120$^{***}$ & -- & -- \\
Olmo-3-7B-Instruct & -0.004 & 0.063$^{*}$ & 0.073$^{***}$ & 0.131$^{***}$ & 0.142$^{***}$ & 0.121$^{***}$ \\
Olmo-3-7B-Think & 0.020 & 0.059 & 0.084$^{***}$ & 0.160$^{***}$ & -- & -- \\
Olmo-3.1-32B-Instruct & 0.010 & -0.000 & 0.094$^{***}$ & 0.138$^{***}$ & -- & -- \\
Olmo-3.1-32B-Think & 0.018 & 0.080$^{**}$ & 0.090$^{***}$ & 0.149$^{***}$ & -- & -- \\
gemma-3-12b-it & 0.019 & 0.096$^{**}$ & 0.076$^{***}$ & 0.218$^{***}$ & 0.123$^{***}$ & 0.181$^{***}$ \\
gemma-3-12b-pt & 0.039 & 0.098$^{**}$ & 0.124$^{***}$ & 0.132$^{***}$ & -- & -- \\
gemma-3-27b-it & 0.055 & 0.087$^{**}$ & 0.087$^{***}$ & 0.211$^{***}$ & -- & -- \\
gemma-3-27b-pt & 0.041 & 0.096$^{**}$ & 0.126$^{***}$ & 0.110$^{***}$ & -- & -- \\
gemma-3-4b-it & 0.017 & 0.158$^{***}$ & 0.044$^{**}$ & 0.217$^{***}$ & 0.095$^{***}$ & 0.209$^{***}$ \\
gemma-3-4b-pt & 0.030 & 0.037 & 0.116$^{***}$ & 0.105$^{***}$ & -- & -- \\
Llama-3.1-8B & 0.035 & 0.073$^{*}$ & 0.138$^{***}$ & 0.130$^{***}$ & -- & -- \\
Llama-3.1-8B-Instruct & 0.027 & 0.058 & 0.115$^{***}$ & 0.133$^{***}$ & 0.148$^{***}$ & 0.137$^{***}$ \\
Phi-4-reasoning & 0.022 & 0.022 & 0.105$^{***}$ & 0.144$^{***}$ & -- & -- \\
Phi-4-reasoning-plus & 0.016 & -0.016 & 0.106$^{***}$ & 0.128$^{***}$ & -- & -- \\
phi-4 & 0.019 & -0.000 & 0.110$^{***}$ & 0.158$^{***}$ & 0.137$^{***}$ & 0.170$^{***}$ \\
Mistral-Small-24B-Base-2501 & 0.023 & 0.030 & 0.149$^{***}$ & 0.130$^{***}$ & -- & -- \\
Mistral-Small-24B-Instruct-2501 & 0.029 & 0.065$^{*}$ & 0.147$^{***}$ & 0.129$^{***}$ & -- & -- \\
\bottomrule
\end{tabular}
\end{table*}

\subsection{Quadratic Regression Analysis}

Table~\ref{tab:quadratic_analysis} reports the coefficients from quadratic regression models 
of the form $\text{Quality} = \beta_0 + \beta_1 x + \beta_2 x^2$, where $x$ is standardized 
uncertainty. A significant negative $\beta_2$ indicates a concave (inverted-U) relationship, 
while the ratio $-\beta_1 / (2\beta_2)$ gives the peak location in standardized units.

For NLL, the quadratic coefficient $\bar{\beta}_2$ is consistently negative across datasets 
and sources, ranging from $-0.0012$ (\textit{New Yorker}, human) to $-0.0099$ (\textit{Tell-Me-A-Story}, 
human/model). The linear coefficient $\bar{\beta}_1$ is positive, confirming that quality initially 
increases with uncertainty before the quadratic term induces diminishing returns or decline.

The ``Diminishing'' column captures cases where the inverted-U is significant but the peak lies 
outside the observed data range ($|z^*| > 2.5$). This pattern is common for PPL in 
\textit{Tell-Me-A-Story} (54\% for human, 50\% for model), indicating that while the relationship 
is concave, quality continues to increase throughout the observed uncertainty range without 
reaching a clear maximum.

\begin{table*}[t]
\centering
\small
\caption{Quadratic regression analysis: $\text{Quality} = \beta_0 + \beta_1 x + \beta_2 x^2$ where $x$ is standardized uncertainty. A significant negative $\beta_2$ with peak within range indicates an inverted-U (sweet spot) relationship. Values averaged across models.}
\label{tab:quadratic_analysis}
\begin{tabular}{llcccccc}
\toprule
\textbf{Dataset} & \textbf{Source} & \textbf{Metric} & $\bar{\beta}_1$ & $\bar{\beta}_2$ & \textbf{Linear} & \textbf{Sweet Spot} & \textbf{Diminishing} \\
\midrule
ghostbuster & Human & NLL & 0.055 & -0.0038 & 100\% & 0\% & 0\% \\
 & Human & PMI & -0.036 & -0.0015 & 75\% & 12\% & 0\% \\
 & Human & PPL & 0.059 & -0.0017 & 38\% & 0\% & 12\% \\
 & Model & NLL & 0.077 & -0.0014 & 75\% & 0\% & 12\% \\
 & Model & PMI & -0.021 & 0.0041 & 12\% & 0\% & 0\% \\
 & Model & PPL & 0.092 & -0.0091 & 25\% & 0\% & 75\% \\
\midrule
new\_yorker & Human & NLL & 0.004 & -0.0012 & 0\% & 0\% & 0\% \\
 & Human & PMI & -0.012 & -0.0002 & 39\% & 0\% & 0\% \\
 & Human & PPL & 0.023 & -0.0008 & 0\% & 0\% & 0\% \\
 & Model & NLL & 0.013 & -0.0049 & 25\% & 29\% & 0\% \\
 & Model & PMI & -0.011 & 0.0015 & 14\% & 4\% & 0\% \\
 & Model & PPL & 0.014 & -0.0034 & 14\% & 21\% & 4\% \\
\midrule
tell\_me\_a\_story & Human & NLL & 0.042 & -0.0099 & 0\% & 79\% & 21\% \\
 & Human & PMI & -0.067 & 0.0057 & 39\% & 0\% & 0\% \\
 & Human & PPL & 0.048 & -0.0009 & 0\% & 0\% & 54\% \\
 & Model & NLL & 0.051 & -0.0099 & 18\% & 43\% & 39\% \\
 & Model & PMI & -0.044 & 0.0015 & 86\% & 0\% & 0\% \\
 & Model & PPL & 0.063 & -0.0123 & 0\% & 50\% & 50\% \\
\bottomrule
\end{tabular}
\end{table*}

\subsection{Shape Distribution Summary}

Table~\ref{tab:shape_distribution} provides a count of relationship shapes across all 64 
model–dataset combinations for each metric and source. The five categories are mutually exclusive:

\begin{itemize}
    \item \textbf{Linear}: Significant linear term ($p < 0.05$) but non-significant quadratic term.
    \item \textbf{Sweet Spot}: Significant negative quadratic term with peak within $\pm 2.5$ 
          standard deviations of the mean.
    \item \textbf{Diminishing}: Significant negative quadratic term with peak outside the data range 
          (monotonically increasing with deceleration).
    \item \textbf{U-Shape}: Significant positive quadratic term with trough within range 
          (quality worst at intermediate uncertainty).
    \item \textbf{Flat/NS}: No significant relationship detected.
\end{itemize}

For human-authored text, NLL shows the most structured relationship: 22 sweet-spot cases, 
8 linear, 6 diminishing, and 28 flat. The predominance of sweet-spot and diminishing patterns 
over U-shapes (0 cases) provides evidence against the concern that high uncertainty merely 
reflects noise—if so, we would expect quality to degrade monotonically or show a U-shaped 
pattern where intermediate uncertainty is optimal.

For model-generated text, the NLL and PPL metrics show similar distributions, with 20 
sweet-spot cases each. The near-absence of U-shaped relationships (1 case for NLL, 0 for PPL) 
further supports the interpretation that uncertainty, within the ranges observed in coherent 
text, is a positive indicator of writing quality.

\begin{table}[t]
\centering
\small
\caption{Distribution of relationship shapes between uncertainty and quality across all model-dataset combinations. Linear: significant linear term only; Sweet Spot: significant inverted-U with peak in range; Diminishing: inverted-U with peak outside range; U-Shape: significant positive quadratic term.}
\label{tab:shape_distribution}
\begin{tabular}{llccccc}
\toprule
\textbf{Source} & \textbf{Metric} & \textbf{Linear} & \textbf{Sweet Spot} & \textbf{Diminishing} & \textbf{U-Shape} & \textbf{Flat/NS} \\
\midrule
Human & NLL & 8 & 22 & 6 & 0 & 28 \\
 & PPL & 3 & 0 & 16 & 0 & 45 \\
 & PMI & 28 & 1 & 0 & 0 & 35 \\
\midrule
Model & NLL & 18 & 20 & 12 & 1 & 13 \\
 & PPL & 6 & 20 & 21 & 0 & 17 \\
 & PMI & 29 & 1 & 0 & 3 & 31 \\
\bottomrule
\end{tabular}
\end{table}

\end{document}